\documentclass{article}

\PassOptionsToPackage{sort&compress}{natbib}

\usepackage{iclr2027_conference}
\setcitestyle{numbers,square,comma}   %

\usepackage{xcolor}   %
\usepackage{graphicx}
\usepackage{amsmath}
\usepackage{amssymb}
\usepackage{amsthm}
\usepackage{tikz}   %
\usetikzlibrary{arrows.meta,positioning,fit,backgrounds,calc}

\usepackage{iftex}
\ifPDFTeX
  \usepackage[T1]{fontenc}
  \usepackage{times}
\else
  \usepackage{fontspec}
\fi

\usepackage{microtype}

\newcommand{\rh}[1]{\noindent\textbf{#1}\hspace{0.5em}}

\usepackage{booktabs}
\usepackage{colortbl}   %
\usepackage{pifont}
\usepackage{algorithm}       %
\usepackage{float}          %
\newcommand{\figlbl}{\fontsize{9}{10.8}\selectfont}
\usepackage{algpseudocode}
\usepackage{wrapfig}

\usepackage{xspace}
\newcommand{\ours}{NitroFusion2\xspace}

\usepackage{url}
\usepackage[pagebackref,breaklinks,colorlinks,allcolors=blue]{hyperref}
\usepackage[capitalize]{cleveref}

\iclrfinalcopy

\title{Bringing Adversarial Loss to the Text Stream\\
of Multimodal Diffusion Transformers}

\author{%
  Dar-Yen Chen$^{1, 2}$\:\; Yi-Zhe Song$^{1}$\\[1mm]
  $^{1}$SketchX, CVSSP, University of Surrey \quad
  $^{2}$Turing Elite Research Labs \\
  \small{\texttt{\{d.chen, y.song\}@surrey.ac.uk}}
}

\begin{document}

\maketitle
\lhead{Preprint.}

\vspace{-1.75em}

\begin{abstract}
\vspace{-3mm}
Adversarial diffusion distillation has conventionally relied on image features for discrimination. We revisit this choice for multimodal diffusion transformers (MMDiTs), whose text stream becomes image-dependent through joint attention. We introduce \emph{Text-Stream Adversarial Distillation} (TSAD), which exploits intermediate text-token states as a new adversarial representation. Because the text stream has already read the image while remaining tied to the text condition, changing that condition changes what information it extracts from the image. Building on this property, we introduce \emph{Aspect-Cued Discrimination} (ACD), which conditions the discriminator on different aspects of the sample, such as typography, objects, or style, to steer which properties are emphasized during adversarial training. Building on the expert-discriminator design of NitroFusion, we integrate TSAD and ACD in \textbf{NitroFusion2}.
NitroFusion2 brings few-step distillation from the conventional 4--8-step regime
down to just 1--2 steps at comparable quality. At one step, it produces sharp rendered text and rich fine-grained details, making one-step generation practical rather than merely a stress test for distillation. At two steps, it remains competitive across the evaluated benchmarks. These results establish the MMDiT text stream as a powerful and steerable representation for adversarial diffusion distillation.
Project page: \href{https://chendaryen.github.io/NitroFusion2.github.io}{chendaryen.github.io/NitroFusion2}.
\end{abstract}

\vspace{-2em}

\begin{figure}[!b]
  \vspace{-2mm}
  \centering
  \includegraphics[width=\linewidth]{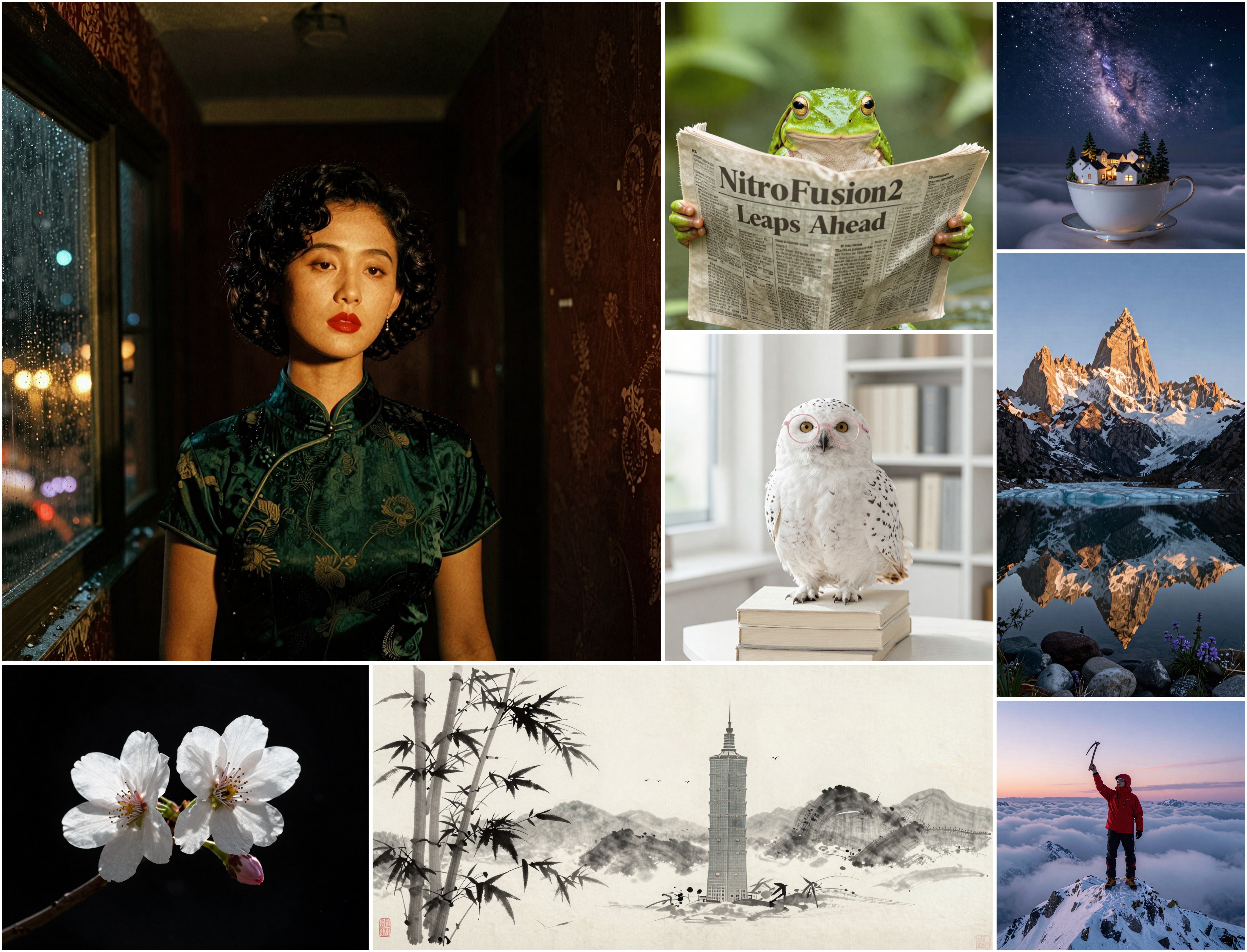}
  \caption{\textbf{Two-step samples from \ours.} Best viewed with zoom.}
  \label{fig:teaser}
\end{figure}

\section{Introduction}
\label{sec:intro}

Adversarial diffusion distillation is an effective approach to few-step generation \cite{yin2024dmd2}: discriminator heads distinguish reference samples from student samples using intermediate features of a pretrained backbone, and the student is trained to fool them \cite{sauer2023add,sauer2024ladd,lin2024lightning,chen2024nitrofusion}. The backbone may be an external encoder or the diffusion model itself. Nearly every part of that recipe has been revisited, from the objectives \cite{yin2024dmd2,xu2025fdistill} and training strategies \cite{ren2024hypersd,cheng2025twinflow} to the discriminator design \cite{sauer2024ladd,chen2024nitrofusion,lin2025apt}, except one: what the discriminator heads read. They read image features. We revisit this choice for multimodal diffusion transformers (MMDiTs).

This choice was natural for earlier diffusion architectures. In a cross-attention UNet \cite{rombach2022ldm,saharia2022imagen,balaji2022ediffi,podell2023sdxl}, the text embeddings used for conditioning remain image-independent throughout the denoising network. The architecture therefore does not maintain text-token hidden states that depend on the sample being processed. MMDiTs \cite{esser2024sd3} change this premise. Through joint attention, intermediate text-token states repeatedly interact with image tokens and become dependent on the particular sample being processed \cite{baek2026omission,ge2026binding,toker2025paddingtone}. The MMDiT text stream therefore exposes a representation that conventional adversarial distillation has largely left unused.

We exploit this representation with \emph{Text-Stream Adversarial Distillation} (TSAD), which attaches lightweight discriminator heads to intermediate text-token states of the discriminator backbone. These heads operate on text-token representations that have already interacted with the image. This turns the text stream itself into a discriminator representation.

In joint attention, text tokens query image features, and what they gather is written back into their own states \cite{li2026registers,toker2025paddingtone,ge2026binding}, so changing the text condition changes what the text stream extracts from the image. If the reference faithfully presents a queried property, such as clear rendered text, the correct object colour, or a particular visual style, the student must reproduce it well enough to fool the discriminator under the same emphasis. Building on this property, we introduce \emph{Aspect-Cued Discrimination} (ACD). For each reference image, we construct an image-grounded structured caption with six fields: typography, object, style, background, action, and layout. At each training step, the discriminator is conditioned on either one valid aspect field or the full structured caption, while the student keeps its original prompt. The cue therefore steers what the text stream queries from the image and which aspects are emphasized by the adversarial signal.

Building on the expert-discriminator design of NitroFusion \cite{chen2024nitrofusion}, we introduce \textbf{NitroFusion2}, distilled from Z-Image-Turbo \cite{team2025zimage}, extending adversarial discrimination to the MMDiT text stream through TSAD and ACD.

Controlled ablations test these two design choices. Text-stream discrimination outperforms conventional image-stream discrimination across all evaluated benchmarks, while combining the two streams performs best on most metrics. Aspect cues further turn the discriminator into an expert on the selected aspect, improving performance on that property. Together, these designs enable a substantial leap in one-step generation. At two steps, \ours{} is competitive with alternatives \cite{liu2026turbopp} that use separate parameter sets for their two-step trajectories.

In summary, our contributions are:
\begin{itemize}

\item We identify the text stream of MMDiTs as a discriminative representation for adversarial diffusion distillation and introduce \emph{Text-Stream Adversarial Distillation}, which applies discriminator heads to intermediate text-token states.

\item We introduce \emph{Aspect-Cued Discrimination}, which uses the discriminator's text conditioning to steer which properties are emphasized by the text stream.

\item We show that text-stream discrimination outperforms conventional image-stream discrimination across all evaluated benchmarks, and that aspect cues yield aspect-specific gains. Together, these advances bring few-step distillation from the conventional
4--8-step regime down to 1--2 steps at comparable quality.

\end{itemize}

If this paper has one message, it is this: in an MMDiT, the text stream is no longer only conditioning. Once it reads the image, it becomes a discriminative representation, opening a new direction for adversarial diffusion distillation and allowing that discrimination to be steered toward specific aspects.

\section{Related Work}
\label{sec:related}

\rh{Few-Step Diffusion Models.}
Reducing diffusion sampling to a few network evaluations has motivated a large
body of distillation work. Progressive Distillation \cite{salimans2022progressive}
pioneered a curriculum that halves the step count iteratively, later generalised
to multistep targets \cite{heek2024multistep}. Consistency models
\cite{song2023cm} and their variants
\cite{song2023ict,luo2023lcm,kim2023ctm,wang2024pcm,ren2024hypersd,lu2024scm,chen2025sanasprint}
enforce self-consistency along the sampling trajectory, mapping any point on it
directly to the endpoint. Distribution Matching Distillation (DMD)
\cite{luo2023diffinstruct,yin2024dmd,yin2024dmd2} minimises a divergence between
student and teacher outputs, and has since been extended in several directions
\cite{xu2025fdistill,liu2026decoupleddmd,jiang2026dmdr,wu2026didr,fan2025phaseddmd,wu2026dpdmd,li2026onexdistill,kim2026dmdnofake,liu2026ctdm,bandyopadhyay2025sd35flash,wang2026advdmd}.
MeanFlow \cite{geng2025meanflow,geng2025imf} learns the average velocity
across an interval. Self-adversarial fine-tuning has the model supply its own
adversarial target \cite{cheng2025twinflow,liu2026apex}. Despite this progress, one- and two-step generation remains challenging \cite{cheng2025twinflow,liu2026turbopp}.

\rh{Adversarial Diffusion Distillation.}
Adversarial objectives recover fidelity lost at low step counts. ADD
\cite{sauer2023add} attaches lightweight discriminator heads to a frozen DINOv2
encoder \cite{oquab2023dinov2} in pixel space. LADD \cite{sauer2024ladd} moves
discrimination into latent space, applying heads to the token sequence the
teacher diffusion model emits after each attention block. NitroFusion
\cite{chen2024nitrofusion} organises the critic as an ensemble of specialised
expert heads in place of a single scorer. Subsequent methods develop these
designs across backbones and head architectures
\cite{xu2023ufogen,kang2024diffusion2gan,lin2024lightning,yin2024dmd2,chen2024nitrofusion,chen2025sanasprint,lin2025apt,wang2026advdmd,bandyopadhyay2025sd35flash},
and several train the critic on teacher generations in place of real images
\cite{sauer2024ladd,chen2024nitrofusion,lin2024lightning,kang2024diffusion2gan,bandyopadhyay2025sd35flash,liu2026turbopp}.
Text-conditioned discrimination was developed earlier in text-to-image GANs
\cite{reed2016gancls,xu2018attngan,miyato2018projection,li2019controlgan,zhang2021xmcgan,tao2022dfgan,sauer2023stylegant}.
GigaGAN \cite{kang2023gigagan} brings it to large-scale models, conditioning
its discriminator on a pooled global descriptor while its generator uses
word-level cross-attention. Across these designs the text conditions the
judgement while the features the discriminator reads come from the image.

\rh{Text-Image Representations in Diffusion Models.}
Diffusion models develop rich text-image correspondences. In cross-attention
UNets, word-level attention maps support attribution \cite{tang2022daam,liu2024understanding},
editing \cite{hertz2022p2p,parmar2023pix2pixzero}, inference-time correction
\cite{chefer2023attendexcite,rassin2023syngen}, and segmentation
\cite{wu2023diffumask,nguyen2023datasetdiffusion}, while the text representation
itself does not vary with the image. Joint attention in multimodal diffusion
transformers \cite{esser2024sd3} changes this. ConceptAttention
\cite{helbling2025conceptattention} localizes concepts in attention outputs,
while recent analyses show that text-stream positions absorb image content
\cite{li2026registers,toker2025paddingtone}, including in image-conditioned
in-context generation \cite{ge2026binding}, causally steer outputs
\cite{li2026textailor}, and support concept prediction from partially denoised
states \cite{baek2026omission}. These representations have been used for
analysis and inference-time control, attention-map supervision
\cite{kim2025seg4diff}, and preference rewards \cite{liu2026dinalrm}, but not
directly as the feature space for adversarial supervision.

\section{Approach}
\label{sec:approach}

\subsection{Preliminaries: Adversarial Diffusion Distillation}
\label{sec:approach:prelim}

A few-step student $G_\theta(x_t,t,p)$ maps a latent $x_t$ at noise level $t$
and original prompt $p$ to a clean latent. For one-step generation,
$\hat{x}=G_\theta(z,1,p)$ with $z\sim\mathcal N(0,I)$. To compare a reference
sample $x$ and a student sample $\hat{x}$ at the same noise level, we use the
flow-matching interpolant \cite{lipman2023flow}
\begin{equation}
\label{eq:interp}
x_t=(1-t)x+t\epsilon,\qquad
\hat{x}_t=(1-t)\hat{x}+t\epsilon,
\quad \epsilon\sim\mathcal N(0,I).
\end{equation}
We denote the resulting reference and student distributions by
$p_{\mathrm{real}}$ and $p_{\mathrm{fake}}$.

Following adversarial diffusion distillation
\cite{sauer2023add,sauer2024ladd}, we reuse a pretrained text-conditioned
diffusion model as a frozen discriminator backbone and attach lightweight
trainable heads to its intermediate states. Writing $D$ for the composition
of the frozen backbone with a trainable head, the adversarial objective
separates reference from student samples,
\begin{equation}
\label{eq:gan}
\mathcal L_{\mathrm{GAN}}
=
\mathbb E_{x_t\sim p_{\mathrm{real}}}
\bigl[\log D(x_t)\bigr]
+
\mathbb E_{\hat{x}_t\sim p_{\mathrm{fake}}}
\bigl[\log\bigl(1-D(\hat{x}_t)\bigr)\bigr],
\end{equation}
and the student is trained to raise the value $D$ assigns to its own samples.

\subsection{Text-Stream Adversarial Distillation}
\label{sec:approach:tsad}

We refer to the evolving image-token and text-token hidden states in an MMDiT
as the image and text streams. Existing adversarial distillation reads
representations from the image stream
\cite{sauer2024ladd,yin2024dmd2,chen2024nitrofusion,lin2025apt,bandyopadhyay2025sd35flash}.
\emph{Text-Stream Adversarial Distillation} (TSAD) instead extends
adversarial discrimination to the text-stream attention outputs.

In a cross-attention UNet \cite{rombach2022ldm}, the text representation
$E(p)$ is computed independently of the image and supplied to the image
network as keys and values. Although the resulting cross-attention weights
depend on image features, the text representation itself remains fixed, with
$\partial E(p)/\partial\hat{x}_t=0$. A discriminator attached directly to
these text states therefore cannot provide an adversarial gradient to the
student.

In an MMDiT \cite{esser2024sd3}, text and image tokens instead interact
through joint attention, whether the two streams carry separate weights or
share them as in Z-Image \cite{team2025zimage}. Consider one layer with
incoming text states
$H_T\in\mathbb R^{m\times d}$. Let $Q_T=H_TW_Q$ denote the text queries, and
let $K=[K_I;K_T]$ and $V=[V_I;V_T]$ collect the image and text keys and
values. The text-query attention weights are
\begin{equation}
\label{eq:attn}
A=
\operatorname{softmax}
\left(
Q_TK^\top/\sqrt d
\right),
\qquad
A=[A_I\;A_T].
\end{equation}
The corresponding attention output is
$U_T=A_IV_I+A_TV_T$, and $U_I$ denotes the image-stream attention output of the
same block, followed by
$H_T'=\psi(H_T+U_TW_O)$, where $\psi$ denotes the remaining token-wise
operations of the block.

The image contribution $A_IV_I$ makes the text attention output directly
dependent on image features \cite{li2026registers,toker2025paddingtone}.
Intuitively, each text token acts as a query into the image: it selects image
information through $A_I$, which contributes directly to its attention output.
This gives sample-dependent text representations whose positions remain
associated with the conditioning text \cite{baek2026omission}. Where image
tokens describe local appearance at fixed spatial positions, the text stream
summarizes the sample along the axes the prompt names, so the adversarial
signal is organized by those axes rather than by local texture alone. TSAD attaches
discriminator heads directly to these attention outputs; image-to-text
attention maps $A_I$ serve for visualization and for the layer selection in
\cref{sec:suppl:layers}.

\Cref{eq:attn} establishes the property required for discrimination, namely
that the text attention outputs depend on the image. \cref{sec:exp} tests
whether they suffice as an adversarial signal.

\begin{figure}[t]
\centering
\includegraphics[width=\linewidth]{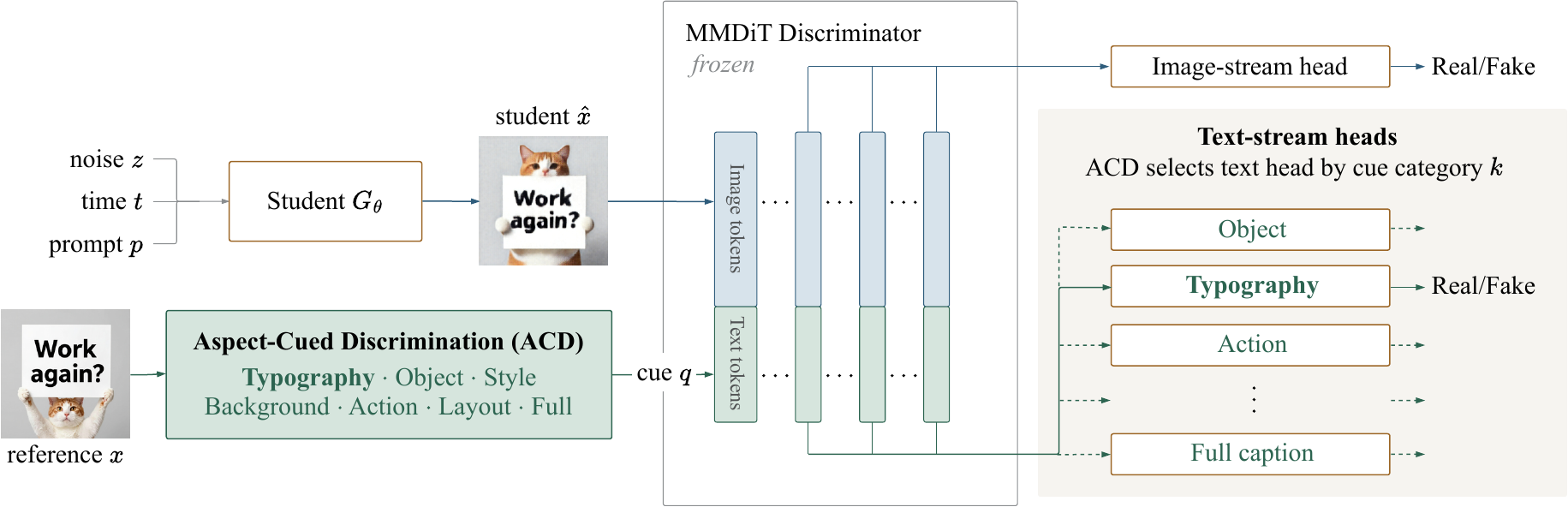}
\caption{\textbf{Architecture of Text-Stream Adversarial Distillation (TSAD) and Aspect-Cued Discrimination (ACD).}
TSAD adds discriminator heads to intermediate text-token states alongside conventional image-stream heads. ACD builds image-grounded aspect fields, then samples one cue per training step. The cue conditions the MMDiT discriminator backbone and selects the corresponding text-stream head, steering discrimination toward that aspect.}
\label{fig:overview}
\end{figure}

As illustrated in \cref{fig:overview}, we attach lightweight discriminator
heads \cite{lin2025apt} to text-stream attention outputs at selected layers of
the frozen MMDiT backbone, with image-stream heads attached to the corresponding
image attention outputs. We denote their logits by
$\ell_q^{\mathrm{text}}(x_t)$ and $\ell_q^{\mathrm{img}}(x_t)$, where $q$ is
the text conditioning supplied to the discriminator backbone. Both sets of
heads read attention outputs from the same backbone pass and provide
adversarial signals to the student.

\subsection{Aspect-Cued Discrimination}
\label{sec:approach:acd}

Because the discriminator reuses a text-conditioned diffusion backbone, its
conditioning prompt can be chosen independently of the original prompt $p$
used by the student. We exploit this flexibility with
\emph{Aspect-Cued Discrimination} (ACD), which conditions the discriminator
on one structured aspect of the sample at a time. Because the backbone is
frozen and the cue enters only as its text input, each cue turns the same
network into a different critic at no additional cost: rather than using a
separate discriminator per property, we re-query one shared backbone for the
property we want to emphasize.

For each reference image, we construct six image-grounded aspect fields indexed
by
\[
\mathcal A=
\{
\mathrm{typography},
\mathrm{object},
\mathrm{style},
\mathrm{background},
\mathrm{action},
\mathrm{layout}
\}.
\]
We denote the text associated with aspect $k$ by $a_k$. Their combination forms
the full structured caption $c_{\mathrm{full}}$. Fields whose content is absent
from the image are left empty. The valid cue categories for a sample are
\[
\mathcal K(a)
=
\{\mathrm{full}\}
\cup
\{k\in\mathcal A:a_k\neq\emptyset\}.
\]
At each training step, we sample
$k\sim\operatorname{Uniform}(\mathcal K(a))$. The discriminator cue $q$ is
$c_{\mathrm{full}}$ when $k=\mathrm{full}$ and $a_k$ otherwise. The student
remains conditioned on the original prompt $p$, while the frozen discriminator
backbone is conditioned on $q$. The same cue is used for the corresponding
reference and student samples.
Following the expert-head design of NitroFusion \cite{chen2024nitrofusion},
the text-stream discriminator uses heads specific to cue categories. A
typography cue is scored by typography heads, an object cue by object heads,
and likewise for the remaining aspects and the full caption. The image-stream
heads remain shared across cues, while their input states still depend on $q$
through the conditioned MMDiT backbone.

The sampled cue therefore affects text-stream discrimination in two coupled
ways. It changes the text queries and hence the image information written into
the text-token states, and it selects the corresponding aspect-specific
text-stream heads. A typography cue, for example, conditions the backbone on
the rendered-text description and routes the resulting states to the typography
heads, while an object cue uses the object description and object heads.

\begin{figure}[t]
\centering
\newlength{\cuepanel}
\setlength{\cuepanel}{0.1215\linewidth}
\newcommand{\cuepic}[1]{\includegraphics[width=\cuepanel]{figures/cueread/#1.pdf}}
\newcommand{\cuehead}[1]{\makebox[\cuepanel]{\figlbl #1}}
{\setlength{\tabcolsep}{0pt}\renewcommand{\arraystretch}{0}%
\begin{tabular}{@{}*{8}{c@{\hspace{1.5pt}}}@{}}
  \rule{0pt}{8pt}\cuehead{Source} & \cuehead{Full caption} &
  \cuehead{Typography} &
  \cuehead{Object} &
  \cuehead{Style} &
  \cuehead{Background} &
  \cuehead{Action} &
  \cuehead{Layout} \\[2pt]
  \cuepic{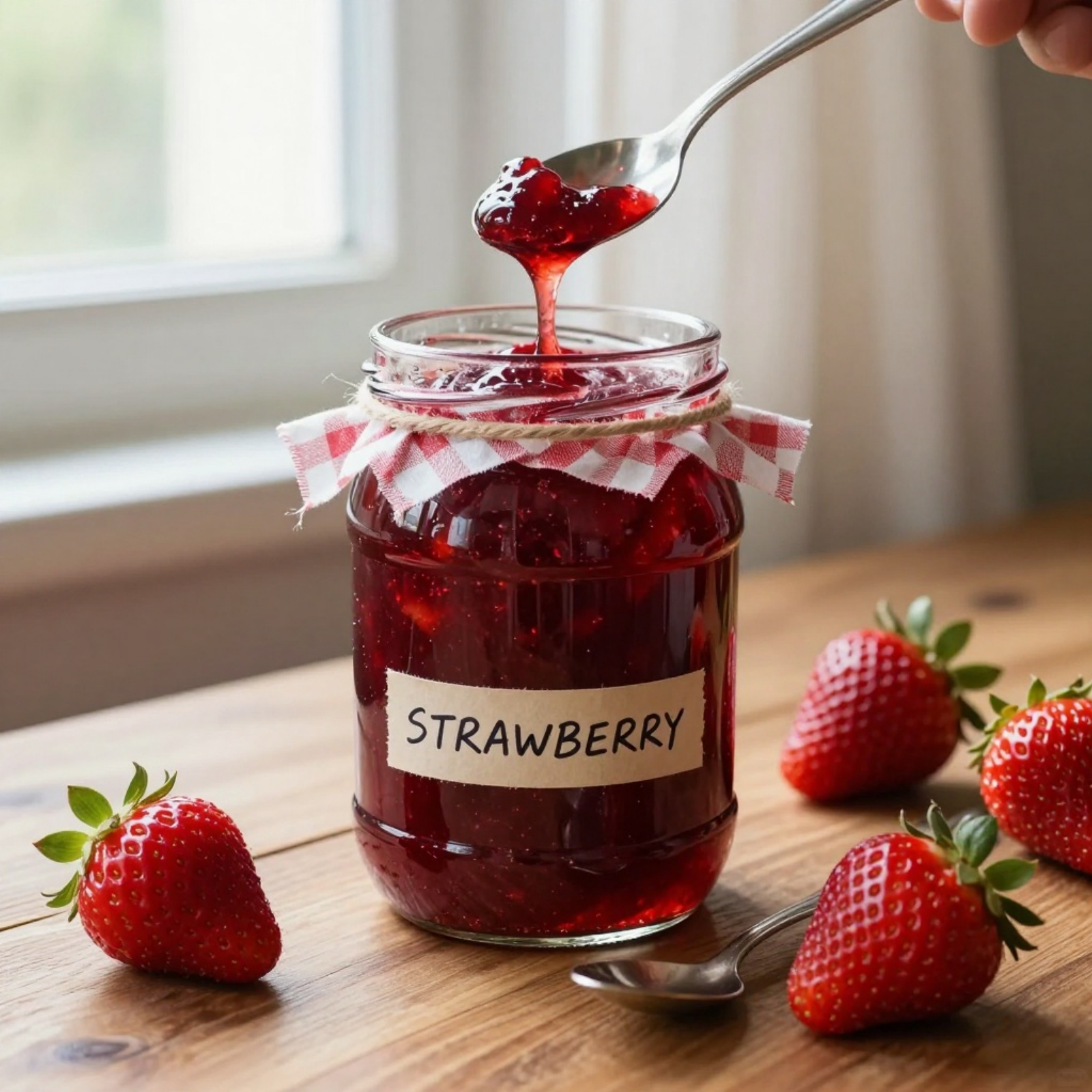} &
  \cuepic{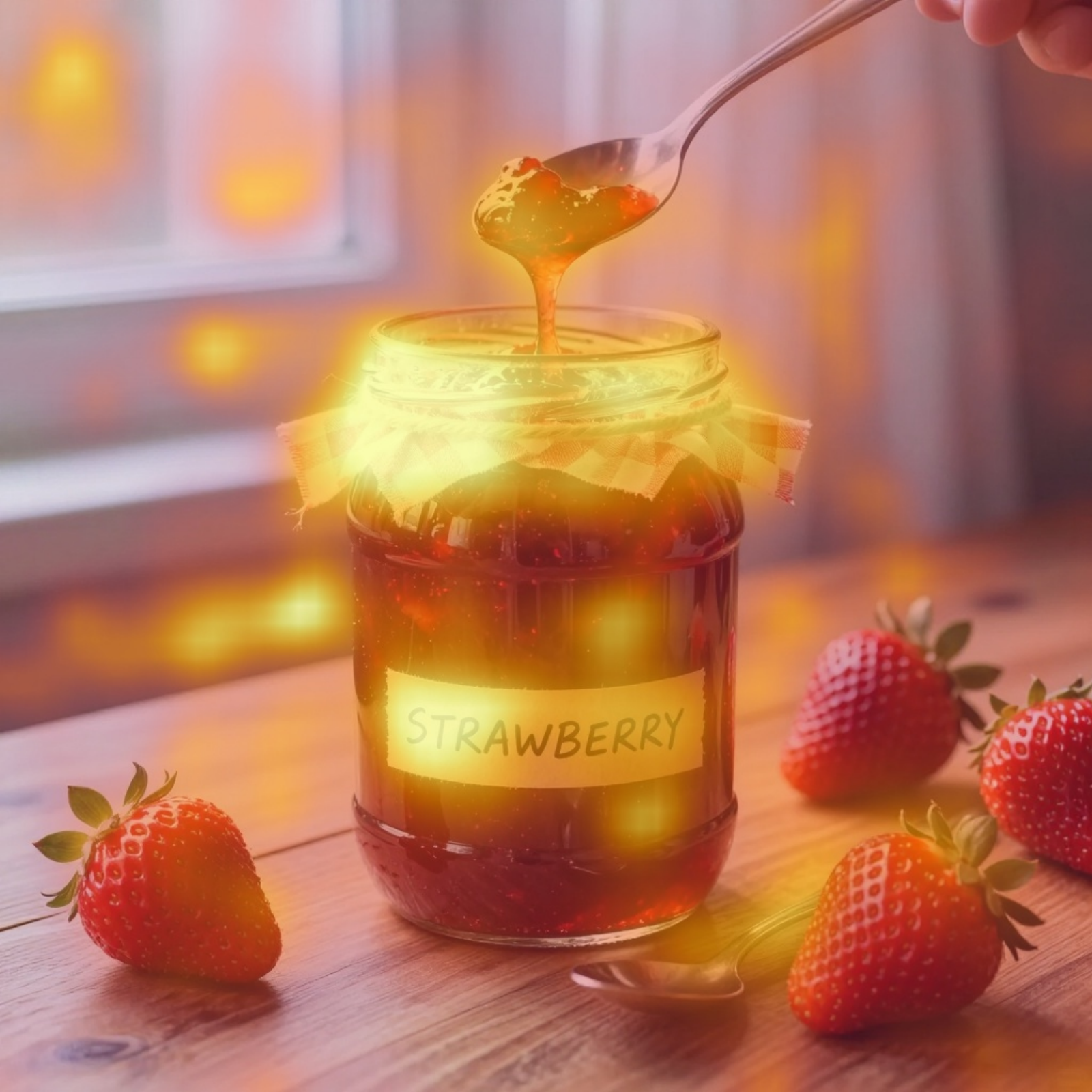} &
  \cuepic{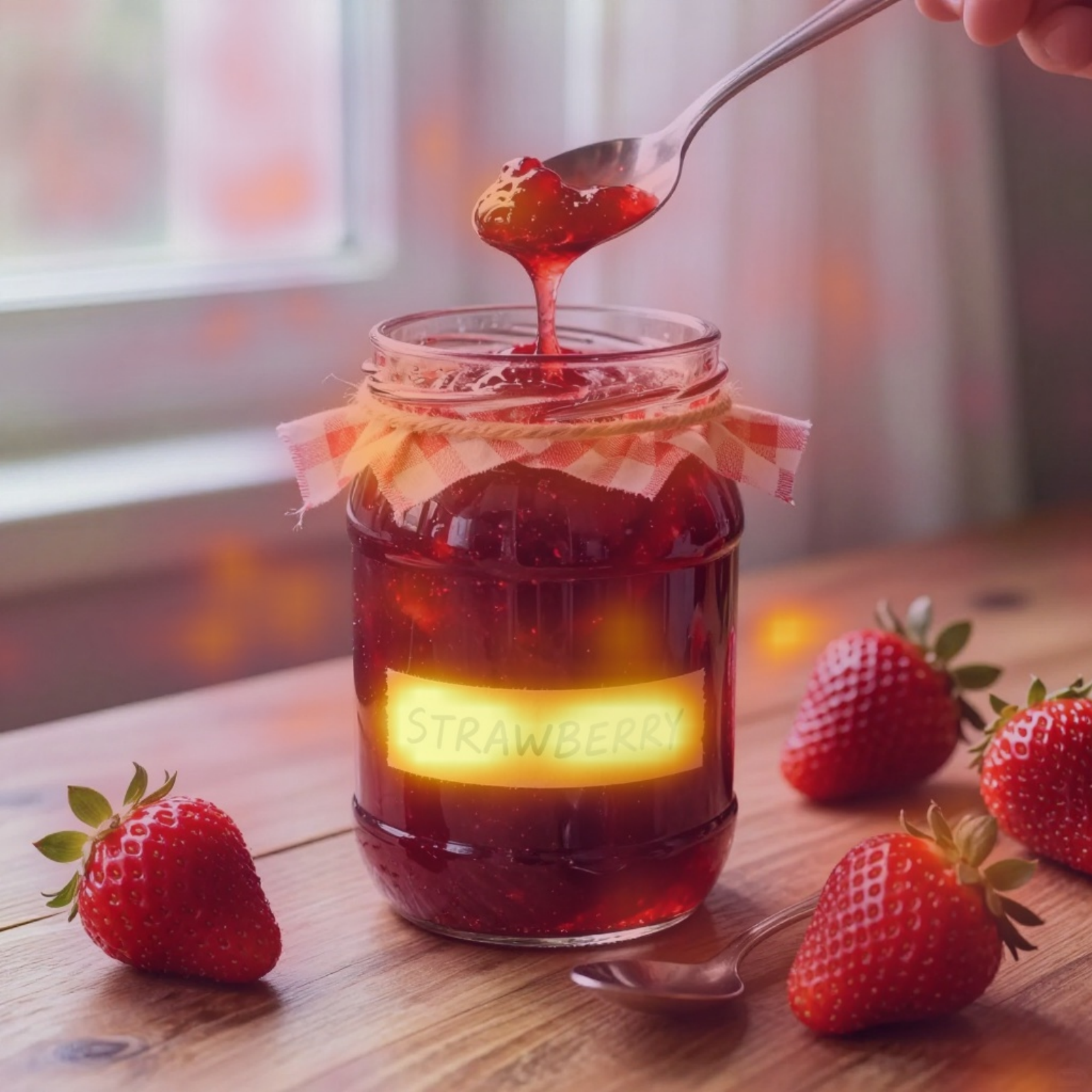} &
  \cuepic{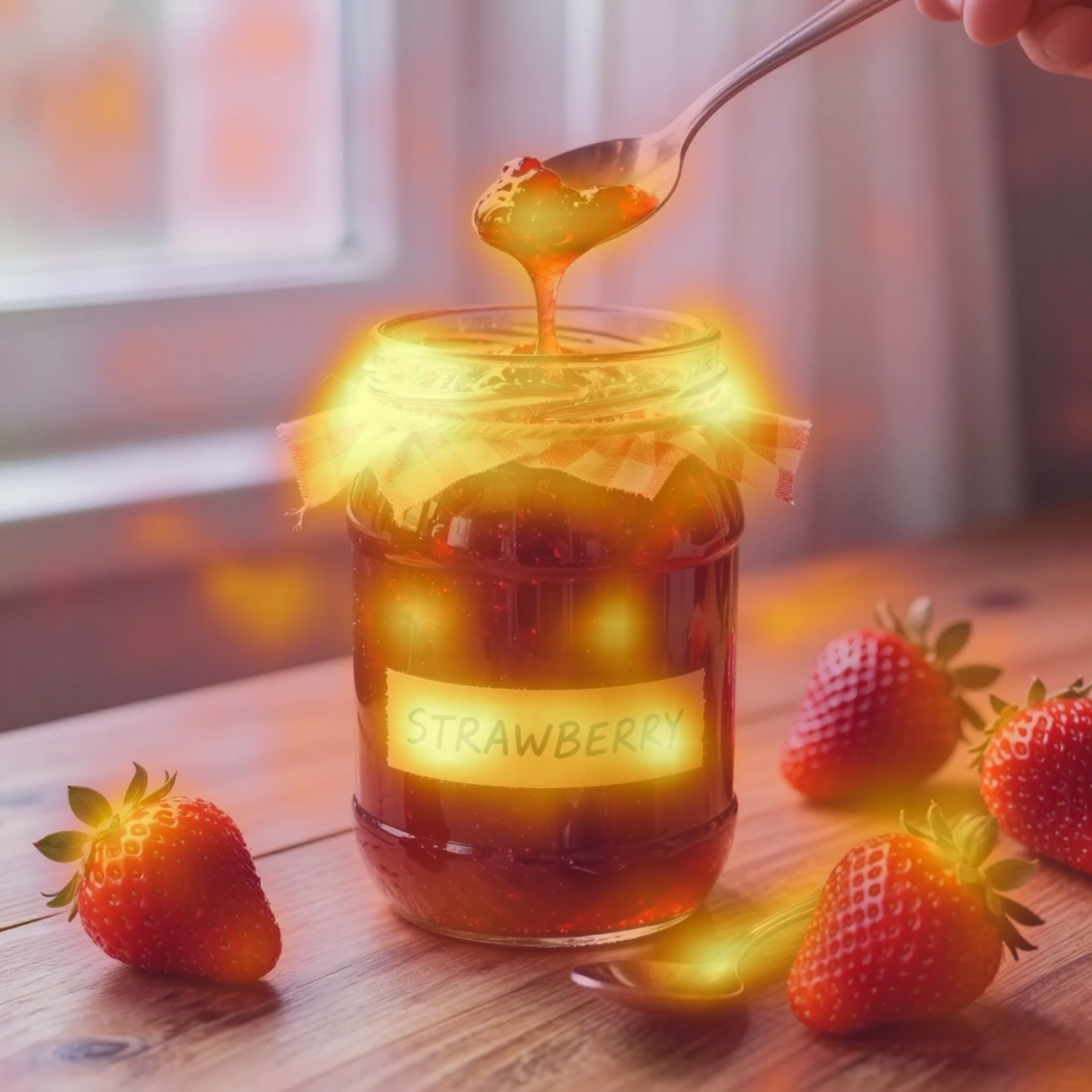} &
  \cuepic{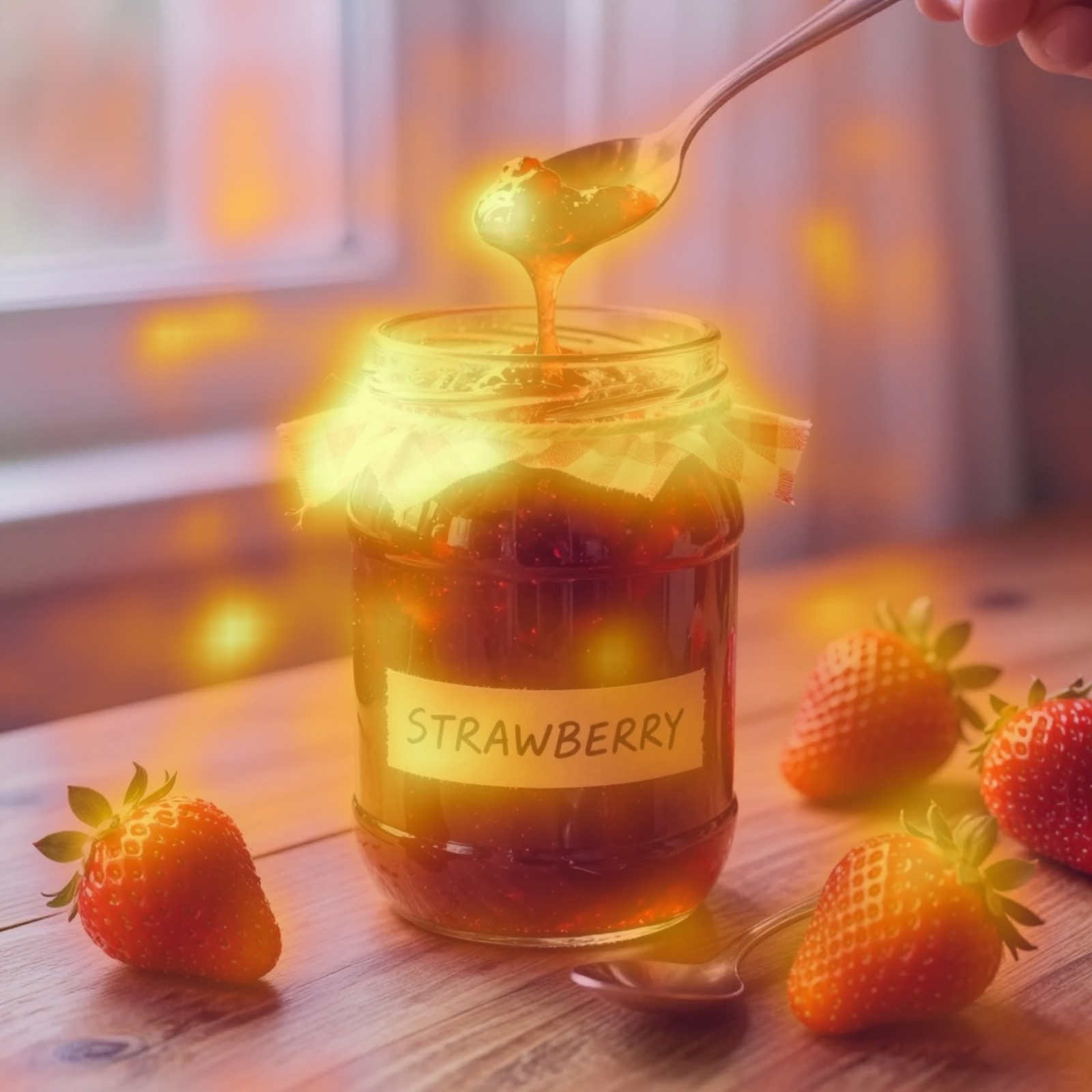} &
  \cuepic{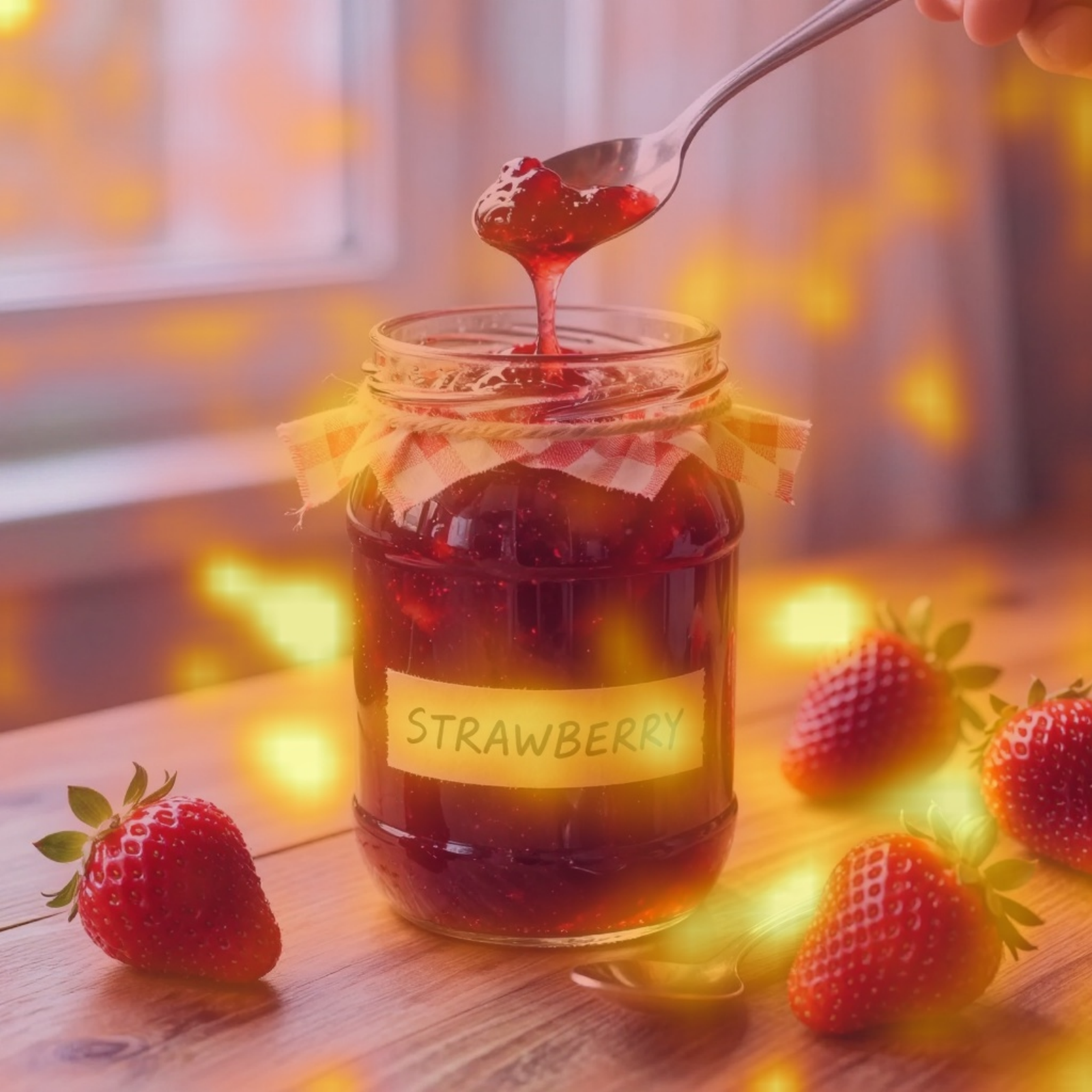} &
  \cuepic{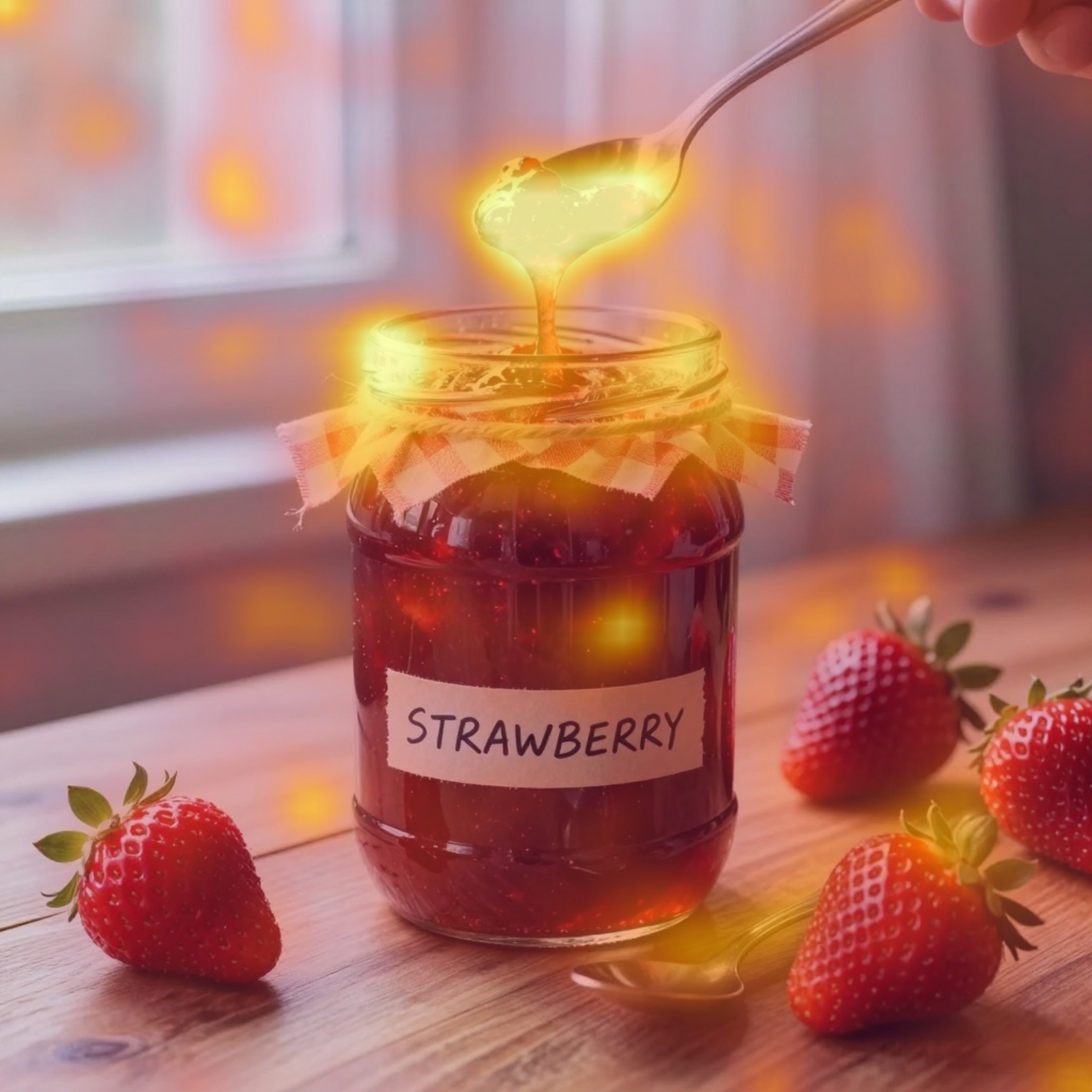} &
  \cuepic{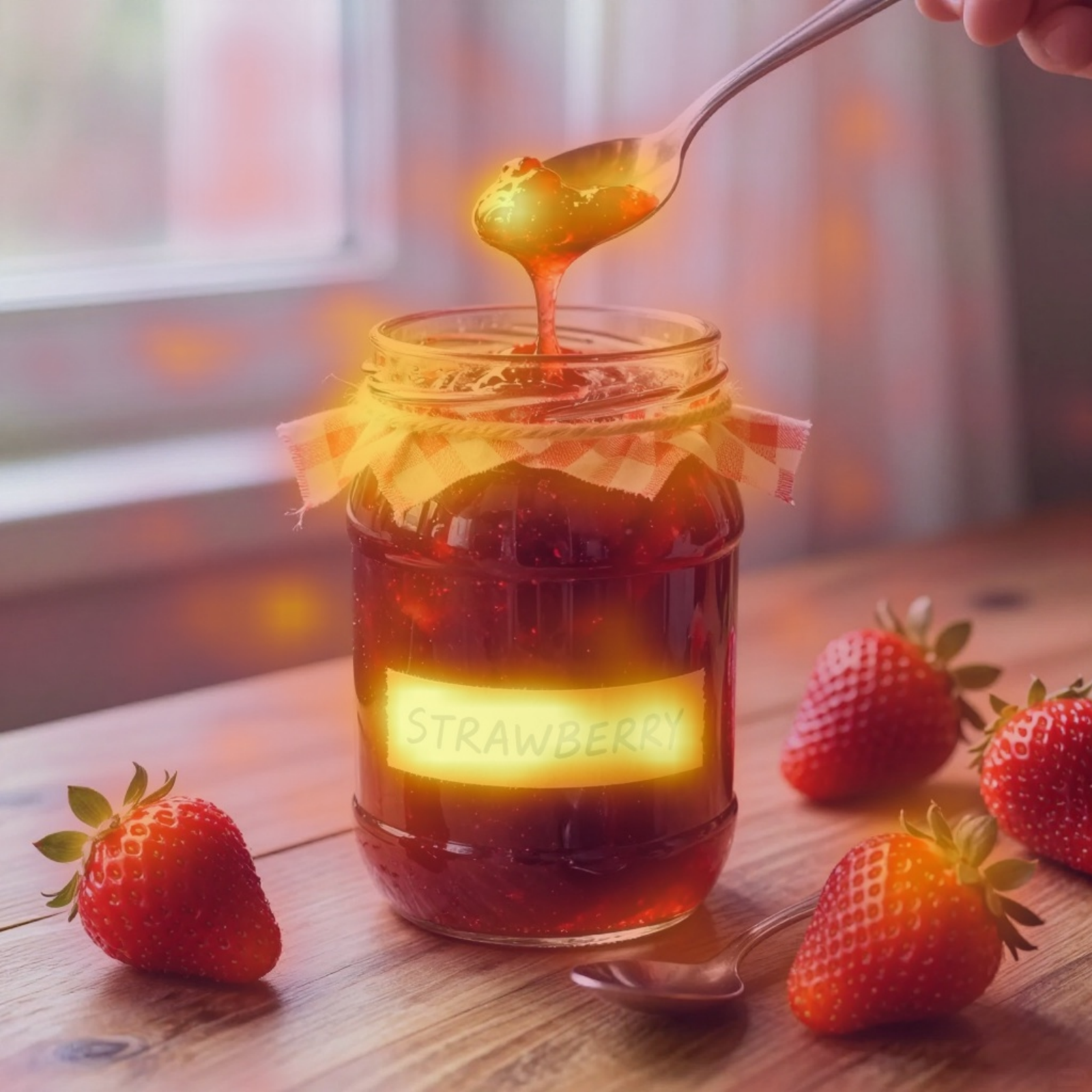} \\
\end{tabular}}

\caption{\textbf{Qualitative cue-dependent image-to-text attention maps.}}
\label{fig:cuemaps}
\end{figure}

\Cref{fig:cuemaps} qualitatively illustrates this cue-dependent attention.
With the same image fixed, different aspect cues produce different
image-to-text attention patterns; for example, the typography and object cues
emphasize different regions associated with rendered text and foreground
objects, the background cue shifts attention away from the foreground toward
the surrounding scene, and the full caption spreads attention more broadly
across the image. These patterns are consistent with ACD steering image-to-text
attention through the discriminator cue. ACD requires no additional backbone
evaluation and only selects the corresponding lightweight text-stream heads.

\subsection{Dataset Construction}
\label{sec:approach:data}

We construct a synthetic dataset of 100k reference images
\cite{sauer2024ladd,chen2024nitrofusion,lin2024lightning,kang2024diffusion2gan,bandyopadhyay2025sd35flash,liu2026turbopp}
using Z-Image-Turbo \cite{team2025zimage} with twelve sampling steps, split
evenly between English and Chinese prompts. Images are generated at a fixed
budget of roughly $1024\times1024$ pixels across a range of aspect ratios. We denote the original prompt by
$p$. For each reference image, the multimodal model Qwen3.5-35B-A3B
\cite{qwen2026qwen35} receives only the image and
generates six image-grounded fields describing typography, object, style,
background, action, and layout. These fields are combined to form the full
structured caption $c_{\mathrm{full}}$.

A field is left empty when the corresponding content is absent from the image.
For example, the typography field is populated only when visible text is
present. We additionally verify that the resulting structured caption is
consistent with the original prompt $p$. During training, the student remains
conditioned on $p$, while the discriminator cue is sampled uniformly from
$c_{\mathrm{full}}$ and the available non-empty aspect fields. The cue
categories are listed in \cref{sec:suppl:cues}.

\begin{figure}[t]
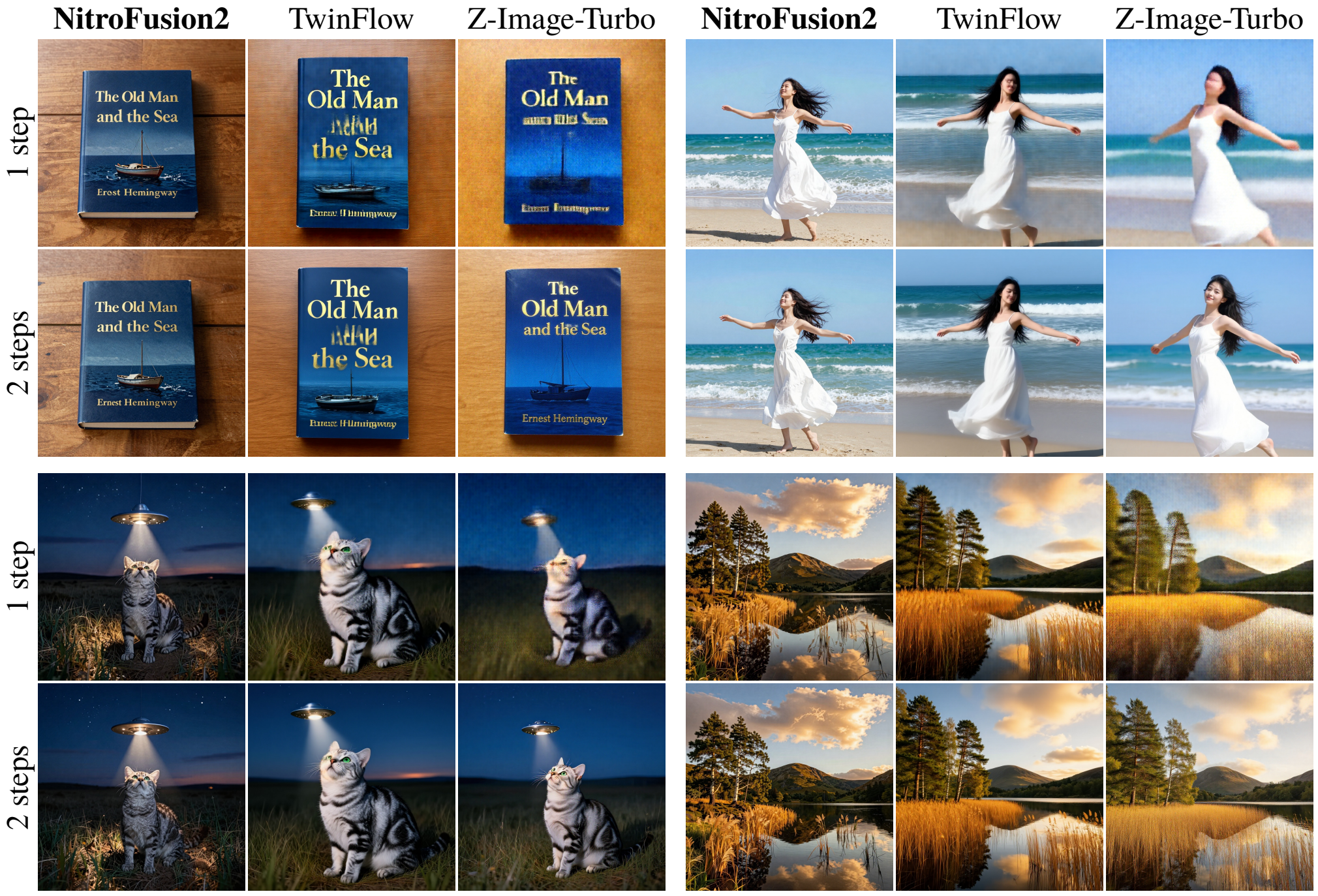

\centering
\newlength{\stepw}\setlength{\stepw}{0.150\linewidth}%
\newcommand{\stepcell}[3]{\includegraphics[width=\stepw]{figures/steps/#1_#2_#3.pdf}}%
\newcommand{\steprow}[3]{%
  \rotatebox{90}{\makebox[\stepw]{\figlbl #3 step\ifnum#3>1 s\fi}} &
  \stepcell{#1}{nf2}{#3} & \stepcell{#1}{twinflow}{#3} & \stepcell{#1}{turbo}{#3} & &
  \stepcell{#2}{nf2}{#3} & \stepcell{#2}{twinflow}{#3} & \stepcell{#2}{turbo}{#3}}%
\newcommand{\stephead}{{\figlbl\textbf{\ours}} & {\figlbl TwinFlow} & {\figlbl Z-Image-Turbo}}%
{\setlength{\tabcolsep}{0.5pt}\renewcommand{\arraystretch}{0}%
\begin{tabular}{@{}c ccc @{\hspace{5pt}} c ccc@{}}
  \rule{0pt}{9pt} & \stephead & & \stephead \\[3pt]
  \steprow{t_book}{s_dancer8}{1} \\[1pt]
  \steprow{t_book}{s_dancer8}{2} \\[5pt]
  \steprow{c_ufo2}{l_tarn}{1} \\[1pt]
  \steprow{c_ufo2}{l_tarn}{2} \\
\end{tabular}}

\caption{\textbf{Qualitative comparison.} \ours preserves rendered text and
fine image detail markedly better than Z-Image-Turbo and TwinFlow.}
\label{fig:compare}
\end{figure}

\begin{figure}[t]
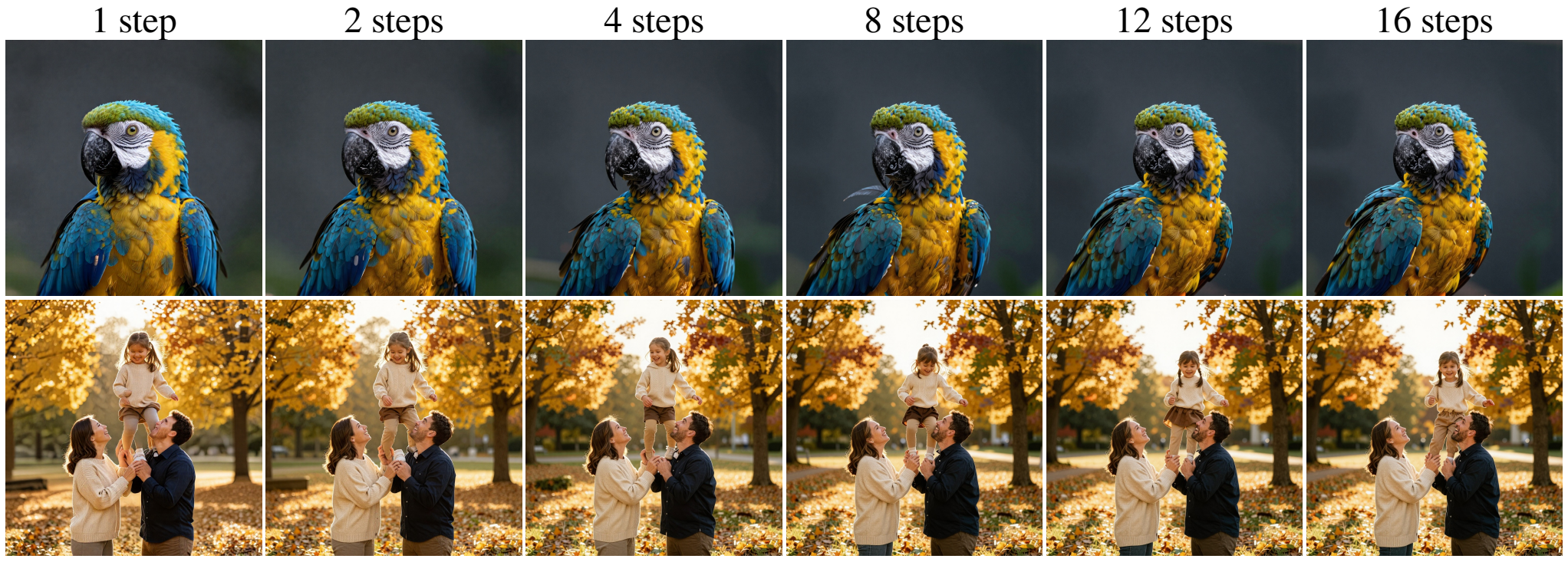

\vspace{-3mm}
\centering
\newlength{\ladderw}\setlength{\ladderw}{0.158\linewidth}%
\newcommand{\laddercell}[2]{\includegraphics[width=\ladderw]{figures/ladder/#1_#2step.pdf}}%
\newcommand{\ladderhead}[1]{\makebox[\ladderw]{\figlbl #1}}%
\newcommand{\ladderrow}[1]{%
  \laddercell{#1}{01} & \laddercell{#1}{02} & \laddercell{#1}{04} &
  \laddercell{#1}{08} & \laddercell{#1}{12} & \laddercell{#1}{16}}%
{\setlength{\tabcolsep}{0.5pt}\renewcommand{\arraystretch}{0}%
\begin{tabular}{@{}*{6}{c}@{}}
  \rule{0pt}{9pt}\ladderhead{1 step} & \ladderhead{2 steps} & \ladderhead{4 steps} &
  \ladderhead{8 steps} & \ladderhead{12 steps} & \ladderhead{16 steps} \\[2pt]
  \ladderrow{parrot} \\[1pt]
  \ladderrow{family} \\
\end{tabular}}

\caption{\textbf{Qualitative results across multiple steps.}}
\label{fig:multistep}
\end{figure}

\begin{table}[t]
\centering
\small
\setlength{\tabcolsep}{4pt}
\caption{\textbf{Few-step students of Z-Image.}}
\label{tab:main}
\begin{tabular}{lcccccccc}
\toprule
& & & \multicolumn{2}{c}{Text rendering} & \multicolumn{3}{c}{Overall} \\
\cmidrule(lr){4-5}\cmidrule(lr){6-8}
Method & Params & NFE & LongText-EN & LongText-CN & GenEval & DPG-Bench & OneIG \\
\midrule
\rowcolor{gray!12}
\textbf{\ours} (ours) & 6B & 1 & \textbf{43.34} & \textbf{47.78} & \textbf{75.86} & \textbf{86.48} & \textbf{47.63} \\
Z-Image-Turbo \cite{team2025zimage} & 6B & 1 & 11.72 & 12.70 & 57.08 & 77.23 & 32.43 \\
TwinFlow \cite{cheng2025twinflow} & 6B & 1 & 29.18 & 36.85 & 66.34 & 85.17 & 42.20 \\
\midrule
\rowcolor{gray!12}
\textbf{\ours} (ours) & 6B & 2 & 76.03 & 81.95 & 74.46 & \textbf{86.21} & 51.45 \\
Z-Image-Turbo \cite{team2025zimage} & 6B & 2 & 72.89 & 78.48 & \textbf{76.53} & 85.78 & 50.94 \\
TwinFlow \cite{cheng2025twinflow} & 6B & 2 & 71.99 & 78.65 & 72.41 & 85.98 & 51.38 \\
Z-Image-Turbo++ \cite{liu2026turbopp} & 6B+6B & 2 & \textbf{89.88} & \textbf{91.62} & 75.70 & 85.86 & \textbf{52.50} \\
\midrule
Z-Image-Turbo \cite{team2025zimage} & 6B & 8 & 91.74 & 92.56 & 75.01 & 84.86 & 52.84 \\
\bottomrule
\end{tabular}

\vspace{2.5mm}
\centering
\small
\setlength{\tabcolsep}{3pt}
\caption{\textbf{Per-task breakdown on GenEval.}}
\label{tab:geneval}
\begin{tabular}{lccccccccc}
\toprule
Method & Params & NFE & Overall & Single & Two & Counting & Colors & Position & Color attr. \\
\midrule
\rowcolor{gray!12}
\textbf{\ours} (ours) & 6B & 1 & \textbf{75.86} & \textbf{100.00} & \textbf{96.97} & \textbf{62.19} & \textbf{91.22} & 32.25 & \textbf{72.50} \\
Z-Image-Turbo \cite{team2025zimage} & 6B & 1 & 57.08 & 97.81 & 51.26 & 55.31 & 76.86 & 31.75 & 29.50 \\
TwinFlow \cite{cheng2025twinflow} & 6B & 1 & 66.34 & 97.50 & 77.02 & 55.31 & 82.45 & \textbf{36.75} & 49.00 \\
\midrule
\rowcolor{gray!12}
\textbf{\ours} (ours) & 6B & 2 & 74.46 & \textbf{100.00} & \textbf{96.46} & 56.88 & \textbf{90.16} & 31.25 & \textbf{72.00} \\
Z-Image-Turbo \cite{team2025zimage} & 6B & 2 & \textbf{76.53} & 99.38 & 89.39 & \textbf{74.38} & 87.77 & \textbf{50.50} & 57.75 \\
TwinFlow \cite{cheng2025twinflow} & 6B & 2 & 72.41 & 99.69 & 87.12 & 67.81 & 85.11 & 40.00 & 54.75 \\
Z-Image-Turbo++ \cite{liu2026turbopp} & 6B+6B & 2 & 75.70 & 99.69 & 87.63 & 69.06 & 84.57 & 48.75 & 64.50 \\
\midrule
Z-Image-Turbo \cite{team2025zimage} & 6B & 8 & 75.01 & 99.69 & 88.89 & 71.88 & 85.11 & 45.75 & 58.75 \\
\bottomrule
\end{tabular}
\end{table}

\subsection{Training Objective}
\label{sec:approach:obj}

We alternate between updating the discriminator heads and the student while
keeping the discriminator backbone frozen. For each stream
$s\in\{\mathrm{text},\mathrm{img}\}$ we use a hinge loss
\cite{lim2017geometric} for adversarial training,
\begin{equation}
\label{eq:disc}
\begin{aligned}
\mathcal L_{\mathrm D}^{s}
=\frac{1}{2}\mathbb E_q\Big[
&\mathbb E_{x_t\sim p_{\mathrm{real}}}
\max\bigl(0,1-\ell_q^s(x_t)\bigr)\\
&+
\mathbb E_{\hat{x}_t\sim p_{\mathrm{fake}}}
\max\bigl(0,1+\ell_q^s(\hat{x}_t)\bigr)
\Big],
\end{aligned}
\end{equation}
where $q$ is sampled according to the ACD procedure in
\cref{sec:approach:acd}.

The student is trained to increase the discriminator logits assigned to its
own samples. The text-stream and image-stream adversarial losses are
\begin{equation}
\label{eq:adv}
\begin{aligned}
\mathcal L_{\mathrm{text-adv}}
&=
-\mathbb E_q
\mathbb E_{\hat{x}_t\sim p_{\mathrm{fake}}}
\left[
\ell_q^{\mathrm{text}}(\hat{x}_t)
\right],\\
\mathcal L_{\mathrm{img-adv}}
&=
-\mathbb E_q
\mathbb E_{\hat{x}_t\sim p_{\mathrm{fake}}}
\left[
\ell_q^{\mathrm{img}}(\hat{x}_t)
\right].
\end{aligned}
\end{equation}

At low noise levels the student is additionally regressed toward the reference
sample,
\begin{equation}
\label{eq:fm}
\mathcal L_{\mathrm{fm}}=
\mathbb E_{t\leq t_{\mathrm{fm}}}
\left[
w(t)
\left\|
G_\theta(x_t,t,p)-x
\right\|_2^2
\right],
\end{equation}
where $t_{\mathrm{fm}}$ specifies the regression range and $w(t)$ weights the
corresponding noise levels. The student objective combines the three terms,
\begin{equation}
\label{eq:genloss}
\mathcal L_{\mathrm G}=
(1-\lambda_{\mathrm{img}})
\mathcal L_{\mathrm{text-adv}}
+
\lambda_{\mathrm{img}}
\mathcal L_{\mathrm{img-adv}}
+
\lambda_{\mathrm{fm}}
\mathcal L_{\mathrm{fm}},
\end{equation}
where $\lambda_{\mathrm{img}}$ controls the balance between text-stream and
image-stream adversarial discrimination. We use
$\lambda_{\mathrm{img}}=0.1$, $\lambda_{\mathrm{fm}}=0.1$ and
$t_{\mathrm{fm}}=0.75$.
\section{Experiments}
\label{sec:exp}

Both our student and our discriminator backbone are initialized from
Z-Image-Turbo \cite{team2025zimage}, and the discriminator backbone is kept
frozen throughout training. We evaluate \ours on LongText-Bench \cite{geng2025xomni} in English and Chinese, OneIG-Bench \cite{chang2025oneig}, GenEval \cite{ghosh2023geneval}, and DPG-Bench \cite{hu2024ella}. We compare against few-step students of Z-Image: Z-Image-Turbo \cite{team2025zimage}, TwinFlow \cite{cheng2025twinflow}, and Z-Image-Turbo++ \cite{liu2026turbopp}. Z-Image-Turbo++ uses step-decoupled weights, with a separate model for each of its two steps. Its weights are not released, so we report its published two-step metrics but cannot evaluate it at one step or include it in our qualitative comparisons. Training details are given in \cref{sec:suppl:impl} and per-dimension benchmark breakdowns in \cref{sec:suppl:extra}. All qualitative samples in this section are generated at $1024\times1024$.

\subsection{Qualitative Results}
\label{sec:exp:qual}

\Cref{fig:compare} compares \ours with Z-Image-Turbo and TwinFlow at one and two
sampling steps. The clearest difference is in rendered text. At one step,
\ours already produces a sharp, legible title, ``The Old Man and the Sea,''
while both baselines blur or distort the lettering. At two steps, \ours also
renders the much smaller author name, ``Ernest Hemingway,'' clearly. Because
the discriminator is conditioned on the described text, the student must
reproduce the glyphs it names, not merely produce sharp lettering.

Beyond typography, \ours already produces realistic, fine-grained details at
one step, including sharp structure in hair, fur, foliage, water, and
reflections. \Cref{fig:multistep} further shows that high-quality generation is
available from the first step and remains stable as the sampling budget
increases. One-step generation can still struggle with complex scenes, but
these errors are progressively alleviated with more steps; in these examples the
visual differences beyond eight steps are small.

\subsection{Quantitative Results}
\label{sec:exp:quant}

\Cref{tab:main} compares few-step students of Z-Image at one and two sampling
steps. At one step, \ours leads all evaluated baselines across all five
benchmarks. Compared with TwinFlow, it substantially improves LongText-EN and
LongText-CN by 14.2 and 10.9 points, respectively, while also improving
GenEval, DPG-Bench, and OneIG. The GenEval breakdown in
\cref{tab:geneval} further shows that this advantage extends across
compositional skills. \ours leads five of six categories, with particularly
strong results on single- and two-object generation, colors, and color
attribution.

At two steps, \ours remains strong overall. It leads four of six GenEval
categories and outperforms TwinFlow on both LongText splits. Z-Image-Turbo++
achieves higher scores on some metrics and uses two step-specific 6B parameter
sets, 12B parameters in total, whereas a single 6B \ours model is used at any
sampling step.

Most notably, \ours leads all evaluated benchmarks at one step. At that single
step it also surpasses the eight-step Z-Image-Turbo reference on DPG-Bench
(86.48 against 84.86) and is on par with it on GenEval (75.86 against 75.01),
at one eighth the sampling cost, while a substantial gap to that reference
remains on text rendering. One step is also where the adversarial signal
matters most, since the student cannot revise its output over later steps and
must place all structure in a single pass.

\subsection{Ablation Study}
\label{sec:exp:abl}

We ablate the two components: first, which stream provides the adversarial representation; second, which cue conditions the shared discriminator backbone. All ablation runs are trained for 1{,}000 steps.

\paragraph{Which stream should the discriminator read?}
\Cref{tab:ablation} compares image-stream, text-stream, and joint discrimination. Text-stream discrimination alone outperforms conventional image-stream discrimination on every benchmark. Combining both streams further improves four of the five metrics, including both LongText splits, DPG-Bench, and OneIG. GenEval is the exception, where the text-only model performs best.

\begin{table}[t]
\centering
\small
\setlength{\tabcolsep}{3.5pt}
\caption{\textbf{Which stream the heads read.} Trained for 1{,}000 steps and
evaluated at 1 NFE.}
\label{tab:ablation}
\begin{tabular}{lccccc}
\toprule
Setting & LongText-EN & LongText-CN & GenEval & DPG-Bench & OneIG \\
\midrule
Text & 22.31 & 24.37 & \textbf{66.71} & 84.39 & 38.45 \\
Image & 17.86 & 19.26 & 66.00 & 82.87 & 35.79 \\
Text + image & \textbf{24.47} & \textbf{26.63} & 64.39 & \textbf{84.86} & \textbf{40.02} \\
\bottomrule
\end{tabular}

\vspace{2.5mm}
\centering
\small
\setlength{\tabcolsep}{3pt}
\caption{\textbf{Which cue conditions the discriminator?} Trained for
1{,}000 steps and evaluated at 1 NFE.}
\label{tab:cuegrid}
\resizebox{\linewidth}{!}{%
\begin{tabular}{lccccccc}
\toprule
& \multicolumn{2}{c}{Text rendering} & Style & \multicolumn{2}{c}{Object} & \multicolumn{2}{c}{Alignment} \\
\cmidrule(lr){2-3}\cmidrule(lr){4-4}\cmidrule(lr){5-6}\cmidrule(lr){7-8}
Cue & LT-EN & LT-CN & OneIG-Style & GenEval-Single & DPG-Entity & OneIG-Align. & DPG-Global \\
\midrule
Typography & \textbf{45.06} & \textbf{53.01} & 32.66 & 99.38 & 92.52 & 78.48 & 79.33 \\
Style & 16.09 & 19.28 & 33.75 & 98.44 & 92.36 & 76.62 & 79.60 \\
Object & 20.82 & 25.19 & 32.67 & \textbf{100.00} & 92.94 & 79.14 & 78.72 \\
Full caption & 17.29 & 19.55 & 30.88 & 99.06 & 92.64 & 76.96 & 79.33 \\
All cues & 24.47 & 26.63 & \textbf{35.79} & 98.44 & \textbf{93.21} & \textbf{79.69} & \textbf{80.85} \\
\bottomrule
\end{tabular}}
\end{table}

These results isolate the contribution of extending adversarial discrimination to the text stream. Text-stream supervision is stronger than image-stream supervision across all five metrics. Our approach retains both because the image-stream signal provides an additional gain on most metrics.

\paragraph{Which cue should condition the discriminator?}
We next fix the joint text- and image-stream discriminator and vary only the cue supplied to its shared backbone. \Cref{tab:cuegrid} compares models trained with a single typography, style, or object cue, the full structured caption, and ACD with all cues sampled uniformly. All settings use the same training data and discriminator architecture. Where \cref{fig:cuemaps} shows that the cue changes where the text stream reads, \cref{tab:cuegrid} asks whether that shows up in what the student learns.

Among the single-aspect models, each cue performs best on the evaluation category associated with the aspect it names. The typography cue gives by far the strongest LongText scores, the style cue gives the strongest OneIG-Style score, and the object cue gives the strongest single-aspect results on both object-focused metrics. The full structured caption does not lead any targeted category. This targeted pattern indicates that changing the cue changes which properties receive the strongest adversarial supervision. Typography alone, for instance, exceeds the final model on LongText, because all of its adversarial supervision is spent on rendered text.

Sampling across all cues produces a broader effect. Rather than specializing
in a single property, the mixed cue policy balances supervision across
different aspects. It does not maximize every targeted category, but gives the
strongest OneIG-Style, DPG-Entity, and both alignment scores in the table.
More importantly, the full ACD mixture outperforms the full structured prompt
on six of seven metrics, supporting the benefit of decomposing the prompt into
specialized aspect cues rather than conditioning on the full caption alone.
In particular, the two alignment metrics are highest under the mixed cue policy
rather than any single-aspect specialization.

\begin{figure}[t]
\centering
\newlength{\ablw}\setlength{\ablw}{0.1315\linewidth}%
\newcommand{\ablcell}[1]{\includegraphics[width=\ablw]{figures/abl/#1.pdf}}%
\newcommand{\ablrow}[1]{\ablcell{r#1l0} & \ablcell{r#1l1} & \ablcell{r#1l2} &
  \ablcell{r#1r0} & \ablcell{r#1r1} & \ablcell{r#1r2} & \ablcell{r#1r3}}%
{\setlength{\tabcolsep}{0.5pt}\renewcommand{\arraystretch}{0}%
\begin{tabular}{@{}ccc @{\hspace{7pt}} cccc@{}}
  {\figlbl Image} & {\figlbl Text} & {\figlbl Text+Image} &
  {\figlbl Typography} & {\figlbl Style} & {\figlbl Object} & {\figlbl All cues} \\[2pt]
  \ablrow{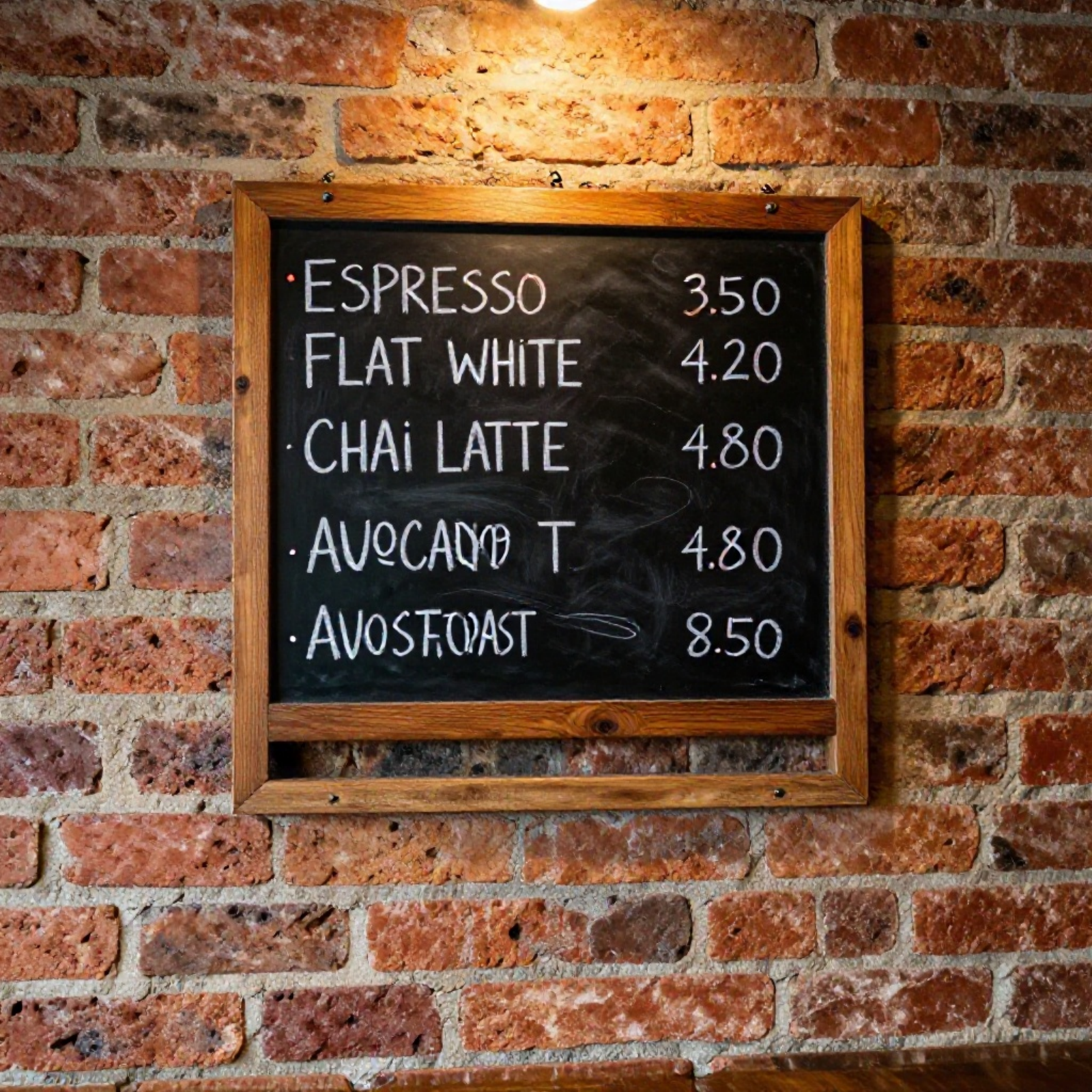} \\[1pt]
  \ablrow{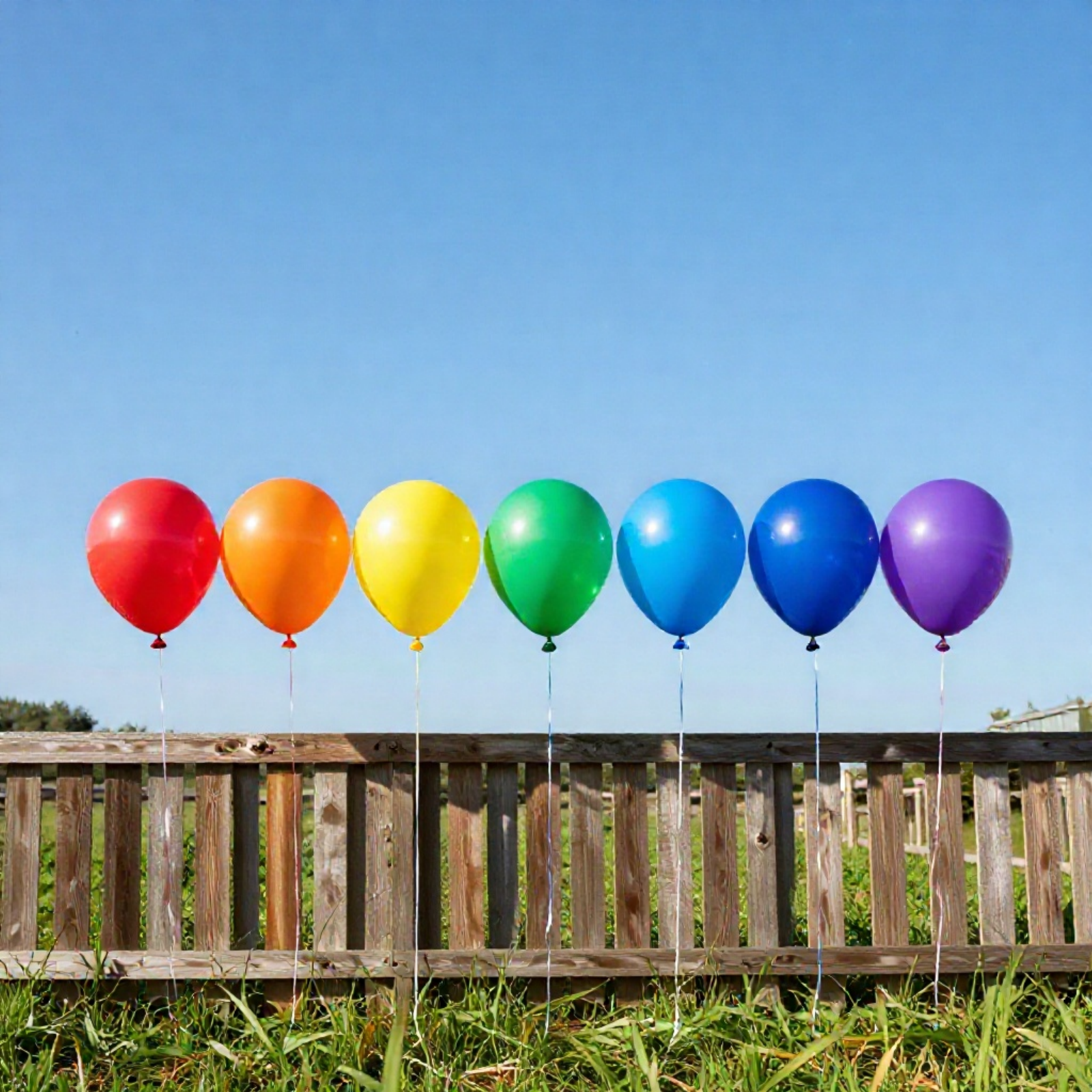}  \\
\end{tabular}}%

\caption{\textbf{Two-step ablative samples.} Left: image-stream, text-stream, and
joint discrimination. Right: models trained with individual typography,
style, or object cues compared with the full seven-cue ACD mixture.}
\label{fig:ablsamples}
\end{figure}

The qualitative ablations in \cref{fig:ablsamples} illustrate the same trends.
On the left, image-stream discrimination struggles with rendered text and
complex structure: it duplicates ``KYOTO'' and introduces artifacts around the
cat, while text-stream and joint discrimination produce cleaner results.
On the right, the effect of different ACD cues is visible across aspects. The
typography cue produces the clearest menu text, while the object cue best
preserves the requested sequence of colored balloons. Sampling across all cues
performs well in both cases, illustrating the broader balance of the full ACD
mixture.

\section{Conclusion and Limitations}
\label{sec:conclusion}

Our experiments show that MMDiT text-token states can serve as an effective
adversarial representation. \textbf{Text-Stream Adversarial Distillation}
(TSAD) exploits these states for supervision, with particularly strong gains
in text rendering, while \textbf{Aspect-Cued Discrimination} (ACD) provides
targeted control and a better overall balance than full-prompt conditioning.
Together, they enable NitroFusion2 to substantially narrow the gap between
one-step and multi-step generation. The text stream is useful here not only
because it carries image information, but because what it carries can be
redirected by the conditioning, making it a steerable representation rather
than a fixed one.

Several limitations remain. Complex scenes are still challenging, particularly
at one step, and sample diversity is reduced. ACD also exhibits a seesaw effect
across cues, where emphasizing one aspect can improve that property while
weakening others. Our layer selection is supported empirically rather than established as optimal.

These observations suggest several directions for future work. Better balancing
the seesaw effect across cues may further improve ACD, while studying where and
how text-stream information evolves across depth could lead to more effective
discriminator designs. Which aspects make effective cues also remains an open
question; the six fields used here are a design choice. More broadly, text-stream states may provide useful
supervision beyond adversarial distillation, opening a wider direction for
training and controlling multimodal diffusion models.

\clearpage
\section*{AI use statement}
In this work, we used generative AI tools and LLMs to generate synthetic data
sets,
implement methods, provide feedback on research methodology and experiments, and
interpret results. We also used them to draft and edit parts of this
paper and to create and modify its figures. We did not use them to propose or
refine the hypotheses of this work or to develop its conceptual framework. We have reviewed all AI-assisted
work: every reported number comes from a run the authors launched and verified.
The authors are responsible for all content in this paper, including anything
produced with AI assistance.

\section*{Reproducibility statement}
The method is defined in \cref{sec:approach:tsad,sec:approach:acd} and the
training objective in \cref{sec:approach:obj}. \Cref{sec:suppl:impl} gives the
training setup. \Cref{sec:suppl:cues} lists the seven cue categories
and \cref{sec:suppl:layers} the layers each set of heads reads.
\Cref{sec:suppl:algo} sets out one full training step as pseudo-code. Dataset
construction, including the annotation model and the six structured-caption
fields, is described in \cref{sec:approach:data}. \ours will be released publicly upon acceptance.

\bibliographystyle{iclr2027_conference}
\bibliography{main}

\begin{thebibliography}{78}
\providecommand{\natexlab}[1]{#1}
\providecommand{\url}[1]{\texttt{#1}}
\expandafter\ifx\csname urlstyle\endcsname\relax
  \providecommand{\doi}[1]{doi: #1}\else
  \providecommand{\doi}{doi: \begingroup \urlstyle{rm}\Url}\fi

\bibitem[Baek et~al.(2026)Baek, Lew, Shin, Lee, and Yoon]{baek2026omission}
Kanghyun Baek, Jaihyun Lew, Chaehun Shin, Jungbeom Lee, and Sungroh Yoon.
\newblock Diagnosing and correcting concept omission in multimodal diffusion
  transformers.
\newblock In \emph{ICML}, 2026.

\bibitem[Balaji et~al.(2022)Balaji, Nah, Huang, Vahdat, Song, Zhang, Kreis,
  Aittala, Aila, Laine, Catanzaro, Karras, and Liu]{balaji2022ediffi}
Yogesh Balaji, Seungjun Nah, Xun Huang, Arash Vahdat, Jiaming Song, Qinsheng
  Zhang, Karsten Kreis, Miika Aittala, Timo Aila, Samuli Laine, Bryan
  Catanzaro, Tero Karras, and Ming-Yu Liu.
\newblock {eDiff-I}: Text-to-image diffusion models with an ensemble of expert
  denoisers.
\newblock \emph{arXiv preprint arXiv:2211.01324}, 2022.

\bibitem[Bandyopadhyay et~al.(2025)Bandyopadhyay, Entezari, Scott, Adithyan,
  Song, and Jampani]{bandyopadhyay2025sd35flash}
Hmrishav Bandyopadhyay, Rahim Entezari, Jim Scott, Reshinth Adithyan, Yi-Zhe
  Song, and Varun Jampani.
\newblock {SD3.5-Flash}: Distribution-guided distillation of generative flows.
\newblock \emph{arXiv preprint arXiv:2509.21318}, 2025.

\bibitem[Chang et~al.(2025)Chang, Fang, Xing, Wu, Cheng, Wang, Zeng, Yu, and
  Chen]{chang2025oneig}
Jingjing Chang, Yixiao Fang, Peng Xing, Shuhan Wu, Wei Cheng, Rui Wang,
  Xianfang Zeng, Gang Yu, and Hai-Bao Chen.
\newblock {OneIG-Bench}: Omni-dimensional nuanced evaluation for image
  generation.
\newblock In \emph{NeurIPS}, 2025.

\bibitem[Chefer et~al.(2023)Chefer, Alaluf, Vinker, Wolf, and
  Cohen-Or]{chefer2023attendexcite}
Hila Chefer, Yuval Alaluf, Yael Vinker, Lior Wolf, and Daniel Cohen-Or.
\newblock Attend-and-excite: Attention-based semantic guidance for
  text-to-image diffusion models.
\newblock In \emph{SIGGRAPH}, 2023.

\bibitem[Chen et~al.(2025{\natexlab{a}})Chen, Bandyopadhyay, Zou, and
  Song]{chen2024nitrofusion}
Dar-Yen Chen, Hmrishav Bandyopadhyay, Kai Zou, and Yi-Zhe Song.
\newblock {NitroFusion}: High-fidelity single-step diffusion through dynamic
  adversarial training.
\newblock In \emph{CVPR}, 2025{\natexlab{a}}.

\bibitem[Chen et~al.(2025{\natexlab{b}})Chen, Xue, Zhao, Yu, Paul, Chen, Cai,
  Han, and Xie]{chen2025sanasprint}
Junsong Chen, Shuchen Xue, Yuyang Zhao, Jincheng Yu, Sayak Paul, Junyu Chen,
  Han Cai, Song Han, and Enze Xie.
\newblock {SANA-Sprint}: One-step diffusion with continuous-time consistency
  distillation.
\newblock In \emph{ICCV}, 2025{\natexlab{b}}.

\bibitem[Cheng et~al.(2026)Cheng, Sun, Li, and Lin]{cheng2025twinflow}
Zhenglin Cheng, Peng Sun, Jianguo Li, and Tao Lin.
\newblock {TwinFlow}: Realizing one-step generation on large models with
  self-adversarial flows.
\newblock In \emph{ICLR}, 2026.

\bibitem[Esser et~al.(2024)Esser, Kulal, Blattmann, Entezari, M{\"u}ller,
  Saini, Levi, Lorenz, Sauer, Boesel, Podell, Dockhorn, English, Lacey,
  Goodwin, Marek, and Rombach]{esser2024sd3}
Patrick Esser, Sumith Kulal, Andreas Blattmann, Rahim Entezari, Jonas
  M{\"u}ller, Harry Saini, Yam Levi, Dominik Lorenz, Axel Sauer, Frederic
  Boesel, Dustin Podell, Tim Dockhorn, Zion English, Kyle Lacey, Alex Goodwin,
  Yannik Marek, and Robin Rombach.
\newblock Scaling rectified flow transformers for high-resolution image
  synthesis.
\newblock In \emph{ICML}, 2024.

\bibitem[Fan et~al.(2026)Fan, Qiu, Wu, Wang, Lin, Ren, Lin, Gong, and
  Yang]{fan2025phaseddmd}
Xiangyu Fan, Zesong Qiu, Zhuguanyu Wu, Fanzhou Wang, Zhiqian Lin, Tianxiang
  Ren, Dahua Lin, Ruihao Gong, and Lei Yang.
\newblock Phased {DMD}: Few-step distribution matching distillation via score
  matching within subintervals.
\newblock In \emph{CVPR}, 2026.

\bibitem[Ge et~al.(2026)Ge, Gandikota, Torralba, and
  Rott~Shaham]{ge2026binding}
Chris Ge, Rohit Gandikota, Antonio Torralba, and Tamar Rott~Shaham.
\newblock Vision-language binding in in-context image generation.
\newblock In \emph{ICML Workshops}, 2026.

\bibitem[Geng et~al.(2025{\natexlab{a}})Geng, Deng, Bai, Kolter, and
  He]{geng2025meanflow}
Zhengyang Geng, Mingyang Deng, Xingjian Bai, J.~Zico Kolter, and Kaiming He.
\newblock Mean flows for one-step generative modeling.
\newblock In \emph{NeurIPS}, 2025{\natexlab{a}}.

\bibitem[Geng et~al.(2026)Geng, Lu, Wu, Shechtman, Kolter, and He]{geng2025imf}
Zhengyang Geng, Yiyang Lu, Zongze Wu, Eli Shechtman, J.~Zico Kolter, and
  Kaiming He.
\newblock Improved mean flows: On the challenges of fastforward generative
  models.
\newblock In \emph{CVPR}, 2026.

\bibitem[Geng et~al.(2025{\natexlab{b}})Geng, Wang, Ma, Li, Rao, Gu, Zhong, Lu,
  Hu, Zhang, Wang, and Jiang]{geng2025xomni}
Zigang Geng, Yibing Wang, Yeyao Ma, Chen Li, Yongming Rao, Shuyang Gu, Zhao
  Zhong, Qinglin Lu, Han Hu, Xiaosong Zhang, Di~Wang, and Jie Jiang.
\newblock {X-Omni}: Reinforcement learning makes discrete autoregressive image
  generative models great again.
\newblock \emph{arXiv preprint arXiv:2507.22058}, 2025{\natexlab{b}}.

\bibitem[Ghosh et~al.(2023)Ghosh, Hajishirzi, and Schmidt]{ghosh2023geneval}
Dhruba Ghosh, Hanna Hajishirzi, and Ludwig Schmidt.
\newblock {GenEval}: An object-focused framework for evaluating text-to-image
  alignment.
\newblock In \emph{NeurIPS}, 2023.

\bibitem[Heek et~al.(2024)Heek, Hoogeboom, and Salimans]{heek2024multistep}
Jonathan Heek, Emiel Hoogeboom, and Tim Salimans.
\newblock Multistep consistency models.
\newblock \emph{arXiv preprint arXiv:2403.06807}, 2024.

\bibitem[Helbling et~al.(2025)Helbling, Meral, Hoover, Yanardag, and
  Chau]{helbling2025conceptattention}
Alec Helbling, Tuna Han~Salih Meral, Ben Hoover, Pinar Yanardag, and Duen~Horng
  Chau.
\newblock {ConceptAttention}: Diffusion transformers learn highly interpretable
  features.
\newblock In \emph{ICML}, 2025.

\bibitem[Hertz et~al.(2023)Hertz, Mokady, Tenenbaum, Aberman, Pritch, and
  Cohen-Or]{hertz2022p2p}
Amir Hertz, Ron Mokady, Jay Tenenbaum, Kfir Aberman, Yael Pritch, and Daniel
  Cohen-Or.
\newblock Prompt-to-prompt image editing with cross attention control.
\newblock In \emph{ICLR}, 2023.

\bibitem[Hu et~al.(2024)Hu, Wang, Fang, Fu, Cheng, and Yu]{hu2024ella}
Xiwei Hu, Rui Wang, Yixiao Fang, Bin Fu, Pei Cheng, and Gang Yu.
\newblock {ELLA}: Equip diffusion models with {LLM} for enhanced semantic
  alignment.
\newblock \emph{arXiv preprint arXiv:2403.05135}, 2024.

\bibitem[Jiang et~al.(2026)Jiang, Liu, Wang, Wu, Li, Li, Jin, Liu, Lu, Li,
  Zhang, Wang, Hoi, Gao, and Yang]{jiang2026dmdr}
Dengyang Jiang, Dongyang Liu, Zanyi Wang, Qilong Wu, Liuzhuozheng Li,
  Hengzhuang Li, Xin Jin, David Liu, Changsheng Lu, Zhen Li, Bo~Zhang, Mengmeng
  Wang, Steven Hoi, Peng Gao, and Harry Yang.
\newblock Distribution matching distillation meets reinforcement learning.
\newblock In \emph{ECCV}, 2026.

\bibitem[Kang et~al.(2023)Kang, Zhu, Zhang, Park, Shechtman, Paris, and
  Park]{kang2023gigagan}
Minguk Kang, Jun-Yan Zhu, Richard Zhang, Jaesik Park, Eli Shechtman, Sylvain
  Paris, and Taesung Park.
\newblock Scaling up {GANs} for text-to-image synthesis.
\newblock In \emph{CVPR}, 2023.

\bibitem[Kang et~al.(2024)Kang, Zhang, Barnes, Paris, Kwak, Park, Shechtman,
  Zhu, and Park]{kang2024diffusion2gan}
Minguk Kang, Richard Zhang, Connelly Barnes, Sylvain Paris, Suha Kwak, Jaesik
  Park, Eli Shechtman, Jun-Yan Zhu, and Taesung Park.
\newblock Distilling diffusion models into conditional {GANs}.
\newblock In \emph{ECCV}, 2024.

\bibitem[Kim et~al.(2025)Kim, Shin, Hong, Yoon, Arnab, Seo, Hong, and
  Kim]{kim2025seg4diff}
Chaehyun Kim, Heeseong Shin, Eunbeen Hong, Heeji Yoon, Anurag Arnab,
  Paul~Hongsuck Seo, Sunghwan Hong, and Seungryong Kim.
\newblock {Seg4Diff}: Unveiling open-vocabulary semantic segmentation in
  text-to-image diffusion transformers.
\newblock In \emph{NeurIPS}, 2025.

\bibitem[Kim et~al.(2024)Kim, Lai, Liao, Murata, Takida, Uesaka, He, Mitsufuji,
  and Ermon]{kim2023ctm}
Dongjun Kim, Chieh-Hsin Lai, Wei-Hsiang Liao, Naoki Murata, Yuhta Takida,
  Toshimitsu Uesaka, Yutong He, Yuki Mitsufuji, and Stefano Ermon.
\newblock Consistency trajectory models: Learning probability flow {ODE}
  trajectory of diffusion.
\newblock In \emph{ICLR}, 2024.

\bibitem[Kim et~al.(2026)Kim, Lee, and Park]{kim2026dmdnofake}
Youngjoong Kim, Deokyeong Lee, and Jaesik Park.
\newblock Distribution matching distillation without fake score network.
\newblock \emph{arXiv preprint arXiv:2605.19256}, 2026.

\bibitem[{Krea AI}(2026)]{krea2turbo}
{Krea AI}.
\newblock {Krea 2 Turbo}.
\newblock \url{https://huggingface.co/krea/Krea-2-Turbo}, 2026.

\bibitem[Li et~al.(2026{\natexlab{a}})Li, Yang, Tan, Zhang, and
  Li]{li2026textailor}
Binglei Li, Mengping Yang, Zhiyu Tan, Junping Zhang, and Hao Li.
\newblock {TexTailor}: Inference-time textual guidance tailoring for multimodal
  diffusion transformers.
\newblock In \emph{ECCV}, 2026{\natexlab{a}}.

\bibitem[Li et~al.(2019)Li, Qi, Lukasiewicz, and Torr]{li2019controlgan}
Bowen Li, Xiaojuan Qi, Thomas Lukasiewicz, and Philip H.~S. Torr.
\newblock Controllable text-to-image generation.
\newblock In \emph{NeurIPS}, 2019.

\bibitem[Li et~al.(2026{\natexlab{b}})Li, Wen, Qi, Wu, Chen, Zhou, Zhu, Wang,
  and Zhang]{li2026onexdistill}
Haoyu Li, Tingyan Wen, Lin Qi, Zhe Wu, Yihuang Chen, Xing Zhou, Lifei Zhu,
  Xueqian Wang, and Kai Zhang.
\newblock 1.x-distill: Breaking the diversity, quality, and efficiency barrier
  in distribution matching distillation.
\newblock \emph{arXiv preprint arXiv:2604.04018}, 2026{\natexlab{b}}.

\bibitem[Li et~al.(2026{\natexlab{c}})Li, Li, Zhou, Li, Chi, Xu, Shen, Xu,
  Tang, Liu, Lan, Qu, and Zhang]{li2026registers}
Maohua Li, Qirui Li, Yanke Zhou, Yiduo Li, Zhaosheng Chi, Chao Xu, Cuifeng
  Shen, Yixuan Xu, Hanlin Tang, Kan Liu, Tao Lan, Lin Qu, and Shao-Qun Zhang.
\newblock Text template tokens are implicit semantic registers in diffusion
  transformers.
\newblock \emph{arXiv preprint arXiv:2607.19139}, 2026{\natexlab{c}}.

\bibitem[Lim \& Ye(2017)Lim and Ye]{lim2017geometric}
Jae~Hyun Lim and Jong~Chul Ye.
\newblock Geometric {GAN}.
\newblock \emph{arXiv preprint arXiv:1705.02894}, 2017.

\bibitem[Lin et~al.(2024)Lin, Wang, and Yang]{lin2024lightning}
Shanchuan Lin, Anran Wang, and Xiao Yang.
\newblock {SDXL-Lightning}: Progressive adversarial diffusion distillation.
\newblock \emph{arXiv preprint arXiv:2402.13929}, 2024.

\bibitem[Lin et~al.(2025)Lin, Xia, Ren, Yang, Xiao, and Jiang]{lin2025apt}
Shanchuan Lin, Xin Xia, Yuxi Ren, Ceyuan Yang, Xuefeng Xiao, and Lu~Jiang.
\newblock Diffusion adversarial post-training for one-step video generation.
\newblock In \emph{ICML}, 2025.

\bibitem[Lipman et~al.(2023)Lipman, Chen, Ben-Hamu, Nickel, and
  Le]{lipman2023flow}
Yaron Lipman, Ricky T.~Q. Chen, Heli Ben-Hamu, Maximilian Nickel, and Matt Le.
\newblock Flow matching for generative modeling.
\newblock In \emph{ICLR}, 2023.

\bibitem[Liu et~al.(2024)Liu, Wang, Cao, Jia, and Huang]{liu2024understanding}
Bingyan Liu, Chengyu Wang, Tingfeng Cao, Kui Jia, and Jun Huang.
\newblock Towards understanding cross and self-attention in stable diffusion
  for text-guided image editing.
\newblock In \emph{CVPR}, 2024.

\bibitem[Liu et~al.(2026{\natexlab{a}})Liu, Sun, Han, Cheng, Chen, and
  Lin]{liu2026apex}
Deyuan Liu, Peng Sun, Yansen Han, Zhenglin Cheng, Chuyan Chen, and Tao Lin.
\newblock Self-adversarial one step generation via condition shifting.
\newblock \emph{arXiv preprint arXiv:2604.12322}, 2026{\natexlab{a}}.

\bibitem[Liu et~al.(2026{\natexlab{b}})Liu, Du, Liu, Jiang, Li, Wu, Li, Hoi,
  Li, and Gao]{liu2026turbopp}
Dongyang Liu, Ruoyi Du, David Liu, Dengyang Jiang, Liangchen Li, Qilong Wu,
  Zhen Li, Steven C.~H. Hoi, Hongsheng Li, and Peng Gao.
\newblock High-fidelity two-step image generation via teacher-aligned
  end-to-end distillation.
\newblock \emph{arXiv preprint arXiv:2606.12575}, 2026{\natexlab{b}}.

\bibitem[Liu et~al.(2026{\natexlab{c}})Liu, Gao, Liu, Du, Li, Wu, Jin, Cao,
  Zhang, Li, and Hoi]{liu2026decoupleddmd}
Dongyang Liu, Peng Gao, David Liu, Ruoyi Du, Zhen Li, Qilong Wu, Xin Jin, Sihan
  Cao, Shifeng Zhang, Hongsheng Li, and Steven Hoi.
\newblock Decoupled {DMD}: {CFG} augmentation as the spear, distribution
  matching as the shield.
\newblock In \emph{ICLR}, 2026{\natexlab{c}}.

\bibitem[Liu et~al.(2026{\natexlab{d}})Liu, Yang, Zhi, Zhong, Ke, Deng, Gao,
  Huang, Zhang, Fu, and Luo]{liu2026dinalrm}
Gongye Liu, Bo~Yang, Yida Zhi, Zhizhou Zhong, Lei Ke, Didan Deng, Han Gao,
  Yongxiang Huang, Kaihao Zhang, Hongbo Fu, and Wenhan Luo.
\newblock Beyond {VLM-Based} rewards: Diffusion-native latent reward modeling.
\newblock In \emph{ICML}, 2026{\natexlab{d}}.

\bibitem[Liu et~al.(2026{\natexlab{e}})Liu, Yan, Chen, Hu, Yue, Pan, Lan, Zhu,
  Cheng, Zheng, and Wang]{liu2026ctdm}
Tao Liu, Hao Yan, Mengting Chen, Taihang Hu, Zhengrong Yue, Zihao Pan, Jinsong
  Lan, Xiaoyong Zhu, Ming-Ming Cheng, Bo~Zheng, and Yaxing Wang.
\newblock Continuous-time distribution matching for few-step diffusion
  distillation.
\newblock \emph{arXiv preprint arXiv:2605.06376}, 2026{\natexlab{e}}.

\bibitem[Lu \& Song(2025)Lu and Song]{lu2024scm}
Cheng Lu and Yang Song.
\newblock Simplifying, stabilizing and scaling continuous-time consistency
  models.
\newblock In \emph{ICLR}, 2025.

\bibitem[Luo et~al.(2023{\natexlab{a}})Luo, Tan, Huang, Li, and
  Zhao]{luo2023lcm}
Simian Luo, Yiqin Tan, Longbo Huang, Jian Li, and Hang Zhao.
\newblock Latent consistency models: Synthesizing high-resolution images with
  few-step inference.
\newblock \emph{arXiv preprint arXiv:2310.04378}, 2023{\natexlab{a}}.

\bibitem[Luo et~al.(2023{\natexlab{b}})Luo, Hu, Zhang, Sun, Li, and
  Zhang]{luo2023diffinstruct}
Weijian Luo, Tianyang Hu, Shifeng Zhang, Jiacheng Sun, Zhenguo Li, and Zhihua
  Zhang.
\newblock Diff-instruct: A universal approach for transferring knowledge from
  pre-trained diffusion models.
\newblock In \emph{NeurIPS}, 2023{\natexlab{b}}.

\bibitem[lvladikov(2026)]{lvladikov2026krea2step}
lvladikov.
\newblock {Krea 2 Turbo} distill 2-step {LoRA}.
\newblock
  \url{https://huggingface.co/lvladikov/Krea2-Turbo-Distill-2step-LoRA}, 2026.

\bibitem[Miyato \& Koyama(2018)Miyato and Koyama]{miyato2018projection}
Takeru Miyato and Masanori Koyama.
\newblock {cGANs} with projection discriminator.
\newblock In \emph{ICLR}, 2018.

\bibitem[Nguyen et~al.(2023)Nguyen, Vu, Tran, and
  Nguyen]{nguyen2023datasetdiffusion}
Quang Nguyen, Truong Vu, Anh Tran, and Khoi Nguyen.
\newblock Dataset diffusion: Diffusion-based synthetic dataset generation for
  pixel-level semantic segmentation.
\newblock In \emph{NeurIPS}, 2023.

\bibitem[Oquab et~al.(2024)Oquab, Darcet, Moutakanni, Vo, Szafraniec, Khalidov,
  Fernandez, Haziza, Massa, El-Nouby, Assran, Ballas, Galuba, Howes, Huang, Li,
  Misra, Rabbat, Sharma, Synnaeve, Xu, Jegou, Mairal, Labatut, Joulin, and
  Bojanowski]{oquab2023dinov2}
Maxime Oquab, Timoth{\'e}e Darcet, Th{\'e}o Moutakanni, Huy Vo, Marc
  Szafraniec, Vasil Khalidov, Pierre Fernandez, Daniel Haziza, Francisco Massa,
  Alaaeldin El-Nouby, Mahmoud Assran, Nicolas Ballas, Wojciech Galuba, Russell
  Howes, Po-Yao Huang, Shang-Wen Li, Ishan Misra, Michael Rabbat, Vasu Sharma,
  Gabriel Synnaeve, Hu~Xu, Herv{\'e} Jegou, Julien Mairal, Patrick Labatut,
  Armand Joulin, and Piotr Bojanowski.
\newblock {DINOv2}: Learning robust visual features without supervision.
\newblock \emph{TMLR}, 2024.

\bibitem[Parmar et~al.(2023)Parmar, Singh, Zhang, Li, Lu, and
  Zhu]{parmar2023pix2pixzero}
Gaurav Parmar, Krishna~Kumar Singh, Richard Zhang, Yijun Li, Jingwan Lu, and
  Jun-Yan Zhu.
\newblock Zero-shot image-to-image translation.
\newblock In \emph{SIGGRAPH}, 2023.

\bibitem[Podell et~al.(2024)Podell, English, Lacey, Blattmann, Dockhorn,
  M{\"u}ller, Penna, and Rombach]{podell2023sdxl}
Dustin Podell, Zion English, Kyle Lacey, Andreas Blattmann, Tim Dockhorn, Jonas
  M{\"u}ller, Joe Penna, and Robin Rombach.
\newblock {SDXL}: Improving latent diffusion models for high-resolution image
  synthesis.
\newblock In \emph{ICLR}, 2024.

\bibitem[{Qwen Team}(2026)]{qwen2026qwen35}
{Qwen Team}.
\newblock {Qwen3.5}: Towards native multimodal agents.
\newblock \url{https://qwen.ai/blog?id=qwen3.5}, 2026.

\bibitem[Rassin et~al.(2023)Rassin, Hirsch, Glickman, Ravfogel, Goldberg, and
  Chechik]{rassin2023syngen}
Royi Rassin, Eran Hirsch, Daniel Glickman, Shauli Ravfogel, Yoav Goldberg, and
  Gal Chechik.
\newblock Linguistic binding in diffusion models: Enhancing attribute
  correspondence through attention map alignment.
\newblock In \emph{NeurIPS}, 2023.

\bibitem[Reed et~al.(2016)Reed, Akata, Yan, Logeswaran, Schiele, and
  Lee]{reed2016gancls}
Scott Reed, Zeynep Akata, Xinchen Yan, Lajanugen Logeswaran, Bernt Schiele, and
  Honglak Lee.
\newblock Generative adversarial text to image synthesis.
\newblock In \emph{ICML}, 2016.

\bibitem[Ren et~al.(2024)Ren, Xia, Lu, Zhang, Wu, Xie, Wang, and
  Xiao]{ren2024hypersd}
Yuxi Ren, Xin Xia, Yanzuo Lu, Jiacheng Zhang, Jie Wu, Pan Xie, Xing Wang, and
  Xuefeng Xiao.
\newblock {Hyper-SD}: Trajectory segmented consistency model for efficient
  image synthesis.
\newblock In \emph{NeurIPS}, 2024.

\bibitem[Rombach et~al.(2022)Rombach, Blattmann, Lorenz, Esser, and
  Ommer]{rombach2022ldm}
Robin Rombach, Andreas Blattmann, Dominik Lorenz, Patrick Esser, and Bj{\"o}rn
  Ommer.
\newblock High-resolution image synthesis with latent diffusion models.
\newblock In \emph{CVPR}, 2022.

\bibitem[Saharia et~al.(2022)Saharia, Chan, Saxena, Li, Whang, Denton,
  Ghasemipour, Ayan, Mahdavi, Lopes, Salimans, Ho, Fleet, and
  Norouzi]{saharia2022imagen}
Chitwan Saharia, William Chan, Saurabh Saxena, Lala Li, Jay Whang, Emily
  Denton, Seyed Kamyar~Seyed Ghasemipour, Burcu~Karagol Ayan, S.~Sara Mahdavi,
  Rapha~Gontijo Lopes, Tim Salimans, Jonathan Ho, David~J. Fleet, and Mohammad
  Norouzi.
\newblock Photorealistic text-to-image diffusion models with deep language
  understanding.
\newblock In \emph{NeurIPS}, 2022.

\bibitem[Salimans \& Ho(2022)Salimans and Ho]{salimans2022progressive}
Tim Salimans and Jonathan Ho.
\newblock Progressive distillation for fast sampling of diffusion models.
\newblock In \emph{ICLR}, 2022.

\bibitem[Sauer et~al.(2023)Sauer, Karras, Laine, Geiger, and
  Aila]{sauer2023stylegant}
Axel Sauer, Tero Karras, Samuli Laine, Andreas Geiger, and Timo Aila.
\newblock {StyleGAN-T}: Unlocking the power of {GANs} for fast large-scale
  text-to-image synthesis.
\newblock In \emph{ICML}, 2023.

\bibitem[Sauer et~al.(2024{\natexlab{a}})Sauer, Boesel, Dockhorn, Blattmann,
  Esser, and Rombach]{sauer2024ladd}
Axel Sauer, Frederic Boesel, Tim Dockhorn, Andreas Blattmann, Patrick Esser,
  and Robin Rombach.
\newblock Fast high-resolution image synthesis with latent adversarial
  diffusion distillation.
\newblock In \emph{SIGGRAPH Asia}, 2024{\natexlab{a}}.

\bibitem[Sauer et~al.(2024{\natexlab{b}})Sauer, Lorenz, Blattmann, and
  Rombach]{sauer2023add}
Axel Sauer, Dominik Lorenz, Andreas Blattmann, and Robin Rombach.
\newblock Adversarial diffusion distillation.
\newblock In \emph{ECCV}, 2024{\natexlab{b}}.

\bibitem[Song \& Dhariwal(2024)Song and Dhariwal]{song2023ict}
Yang Song and Prafulla Dhariwal.
\newblock Improved techniques for training consistency models.
\newblock In \emph{ICLR}, 2024.

\bibitem[Song et~al.(2023)Song, Dhariwal, Chen, and Sutskever]{song2023cm}
Yang Song, Prafulla Dhariwal, Mark Chen, and Ilya Sutskever.
\newblock Consistency models.
\newblock In \emph{ICML}, 2023.

\bibitem[Tang et~al.(2023)Tang, Liu, Pandey, Jiang, Yang, Kumar, Stenetorp,
  Lin, and Ture]{tang2022daam}
Raphael Tang, Linqing Liu, Akshat Pandey, Zhiying Jiang, Gefei Yang, Karun
  Kumar, Pontus Stenetorp, Jimmy Lin, and Ferhan Ture.
\newblock What the {DAAM}: Interpreting stable diffusion using cross attention.
\newblock In \emph{ACL}, 2023.

\bibitem[Tao et~al.(2022)Tao, Tang, Wu, Jing, Bao, and Xu]{tao2022dfgan}
Ming Tao, Hao Tang, Fei Wu, Xiao-Yuan Jing, Bing-Kun Bao, and Changsheng Xu.
\newblock {DF-GAN}: A simple and effective baseline for text-to-image
  synthesis.
\newblock In \emph{CVPR}, 2022.

\bibitem[Toker et~al.(2025)Toker, Galil, Orgad, Gal, Tewel, Chechik, and
  Belinkov]{toker2025paddingtone}
Michael Toker, Ido Galil, Hadas Orgad, Rinon Gal, Yoad Tewel, Gal Chechik, and
  Yonatan Belinkov.
\newblock Padding tone: A mechanistic analysis of padding tokens in {T2I}
  models.
\newblock In \emph{NAACL}, 2025.

\bibitem[Wang et~al.(2024)Wang, Huang, Bergman, Shen, Gao, Lingelbach, Sun,
  Bian, Song, Liu, Wang, and Li]{wang2024pcm}
Fu-Yun Wang, Zhaoyang Huang, Alexander~William Bergman, Dazhong Shen, Peng Gao,
  Michael Lingelbach, Keqiang Sun, Weikang Bian, Guanglu Song, Yu~Liu, Xiaogang
  Wang, and Hongsheng Li.
\newblock Phased consistency models.
\newblock In \emph{NeurIPS}, 2024.

\bibitem[Wang et~al.(2026)Wang, Li, Gong, Ge, and Deng]{wang2026advdmd}
Xu~Wang, Zexian Li, Litong Gong, Tiezheng Ge, and Zhijie Deng.
\newblock {AdvDMD}: Adversarial reward meets {DMD} for high-quality few-step
  generation.
\newblock \emph{arXiv preprint arXiv:2604.28126}, 2026.

\bibitem[Wu et~al.(2025)Wu, Li, Zhou, Lin, Gao, and Yan]{wu2025qwenimage}
Chenfei Wu, Jiahao Li, Jingren Zhou, Junyang Lin, Kaiyuan Gao, and Kun Yan.
\newblock Qwen-image technical report.
\newblock \emph{arXiv preprint arXiv:2508.02324}, 2025.

\bibitem[Wu et~al.(2026{\natexlab{a}})Wu, Luo, Zheng, Zhang, and
  Lin]{wu2026didr}
Junyi Wu, Weijian Luo, Haoyang Zheng, Ruizhe Zhang, and Guang Lin.
\newblock Diff-instruct with diffused reward: Towards principled one-step
  generator {RL}.
\newblock \emph{arXiv preprint arXiv:2605.24001}, 2026{\natexlab{a}}.

\bibitem[Wu et~al.(2026{\natexlab{b}})Wu, Li, Zhang, and Ma]{wu2026dpdmd}
Tianhe Wu, Ruibin Li, Lei Zhang, and Kede Ma.
\newblock Diversity-preserved distribution matching distillation for fast
  visual synthesis.
\newblock In \emph{ICML}, 2026{\natexlab{b}}.

\bibitem[Wu et~al.(2023)Wu, Zhao, Shou, Zhou, and Shen]{wu2023diffumask}
Weijia Wu, Yuzhong Zhao, Mike~Zheng Shou, Hong Zhou, and Chunhua Shen.
\newblock {DiffuMask}: Synthesizing images with pixel-level annotations for
  semantic segmentation using diffusion models.
\newblock In \emph{ICCV}, 2023.

\bibitem[{Wuli-art}(2026)]{wuli2026qwen2step}
{Wuli-art}.
\newblock {Qwen-Image-2512} turbo {LoRA}, 2 steps.
\newblock
  \url{https://huggingface.co/Wuli-art/Qwen-Image-2512-Turbo-LoRA-2-Steps},
  2026.

\bibitem[Xu et~al.(2018)Xu, Zhang, Huang, Zhang, Gan, Huang, and
  He]{xu2018attngan}
Tao Xu, Pengchuan Zhang, Qiuyuan Huang, Han Zhang, Zhe Gan, Xiaolei Huang, and
  Xiaodong He.
\newblock {AttnGAN}: Fine-grained text to image generation with attentional
  generative adversarial networks.
\newblock In \emph{CVPR}, 2018.

\bibitem[Xu et~al.(2024)Xu, Zhao, Xiao, and Hou]{xu2023ufogen}
Yanwu Xu, Yang Zhao, Zhisheng Xiao, and Tingbo Hou.
\newblock {UFOGen}: You forward once large scale text-to-image generation via
  diffusion {GANs}.
\newblock In \emph{CVPR}, 2024.

\bibitem[Xu et~al.(2025)Xu, Nie, and Vahdat]{xu2025fdistill}
Yilun Xu, Weili Nie, and Arash Vahdat.
\newblock One-step diffusion models with $f$-divergence distribution matching.
\newblock \emph{arXiv preprint arXiv:2502.15681}, 2025.

\bibitem[Yin et~al.(2024{\natexlab{a}})Yin, Gharbi, Park, Zhang, Shechtman,
  Durand, and Freeman]{yin2024dmd2}
Tianwei Yin, Micha{\"e}l Gharbi, Taesung Park, Richard Zhang, Eli Shechtman,
  Fredo Durand, and William~T. Freeman.
\newblock Improved distribution matching distillation for fast image synthesis.
\newblock In \emph{NeurIPS}, 2024{\natexlab{a}}.

\bibitem[Yin et~al.(2024{\natexlab{b}})Yin, Gharbi, Zhang, Shechtman, Durand,
  Freeman, and Park]{yin2024dmd}
Tianwei Yin, Micha{\"e}l Gharbi, Richard Zhang, Eli Shechtman, Fredo Durand,
  William~T. Freeman, and Taesung Park.
\newblock One-step diffusion with distribution matching distillation.
\newblock In \emph{CVPR}, 2024{\natexlab{b}}.

\bibitem[{Z-Image Team} et~al.(2025){Z-Image Team}, Cai, Cao, Du, Gao, and
  Hao]{team2025zimage}
{Z-Image Team}, Huanqia Cai, Sihan Cao, Ruoyi Du, Peng Gao, and Aiming Hao.
\newblock Z-image: An efficient image generation foundation model with
  single-stream diffusion transformer.
\newblock \emph{arXiv preprint arXiv:2511.22699}, 2025.

\bibitem[Zhang et~al.(2021)Zhang, Koh, Baldridge, Lee, and
  Yang]{zhang2021xmcgan}
Han Zhang, Jing~Yu Koh, Jason Baldridge, Honglak Lee, and Yinfei Yang.
\newblock Cross-modal contrastive learning for text-to-image generation.
\newblock In \emph{CVPR}, 2021.

\end{thebibliography}

\clearpage
\appendix
\section{Implementation Details}
\label{sec:suppl:impl}

We use Adam with $\beta_1=0.0$, $\beta_2=0.9$, no
weight decay, a learning rate of $5\times10^{-5}$ and a batch size of 256, on
eight H100 GPUs. The final model is trained for 20{,}000 steps in two stages. The first
5{,}000 steps are off-policy. The remaining 15{,}000 are on-policy, where the
student's input is a state rolled out along its own trajectory, and we set
$\lambda_{\mathrm{fm}}=0$, dropping the flow-matching term.

The discriminator heads follow the design of APT \citep{lin2025apt}.
Following NitroFusion \citep{chen2024nitrofusion}, the image stream and each
cue carry a pool of three heads rather than a single head, and one head is
drawn at random per step. We also refresh the heads periodically by
re-initializing their weights, which reduces overfitting.

\subsection{The Cue List}
\label{sec:suppl:cues}

\Cref{tab:cues} lists the seven cue categories and the text each one supplies
to the discriminator. The structured caption they are drawn from is written by
Qwen3.5-35B-A3B \citep{qwen2026qwen35} from the reference
image itself, so each field is detailed and closely aligned with that image.

\begin{table}[h]
\centering
\footnotesize
\setlength{\tabcolsep}{4pt}
\renewcommand{\arraystretch}{1.15}
\caption{\textbf{The seven cue categories.}}
\label{tab:cues}
\begin{tabular}{@{}lp{9.6cm}@{}}
\toprule
Cue & Text the discriminator is conditioned on \\
\midrule
Full caption & Every field concatenated: the complete specification of
the image. \\
\addlinespace[3pt]
Typography & The rendered text, transcribed line by line with each line's
position and lettering. Populated only when the image contains visible text. \\
\addlinespace[3pt]
Object & The subject: what is depicted, with the counts and identities of the
entities preserved. \\
\addlinespace[3pt]
Style & The palette, artistic medium and rendering treatment, describing how
the image is rendered rather than what it shows. \\
\addlinespace[3pt]
Background & The setting the subject sits in: surroundings, surfaces, weather
and ambient detail, excluding the subject itself. \\
\addlinespace[3pt]
Action & What the subject is doing, or an explicit statement that the scene is
static. \\
\addlinespace[3pt]
Layout & Where elements sit in the frame and how they are placed relative to one
another. \\
\bottomrule
\end{tabular}
\end{table}

\subsection{Choosing Where the Heads Read}
\label{sec:suppl:layers}
We score each layer of the frozen backbone by how far a cue's attention map
departs from the full caption's. The full caption itself is scored by its mean
departure from the other six cues. Cues separate more after layer 20, so the
four best-scoring layers all fall in one band. We keep the two strongest and
replace the others with the best-separating layer of the mid and of the early
range, so a head also reads coarser shared structure (\cref{tab:layers}). The image stream gets no such selection: its
states are not indexed by the cue, so there is no per-cue attention map to rank.
We spread its taps evenly across depth instead, at layers 5, 13, 21 and 29.

\begin{table}[!h]
\centering
\footnotesize
\setlength{\tabcolsep}{7pt}
\renewcommand{\arraystretch}{1.15}
\caption{\textbf{Read layers per cue.} Indices into the backbone's 30
joint-attention layers.}
\label{tab:layers}
\begin{tabular}{@{}lcc@{}}
\toprule
Cue & Top four, best first & Selected \\
\midrule
Full caption & $[22, 21, 23, 20]$ & $[22, 21, 17, 10]$ \\
Typography & $[22, 21, 20, 24]$ & $[22, 21, 17, 10]$ \\
Object & $[21, 20, 22, 24]$ & $[21, 20, 17, 10]$ \\
Style & $[21, 22, 20, 25]$ & $[21, 22, 17, 10]$ \\
Background & $[21, 22, 20, 23]$ & $[21, 22, 17, 10]$ \\
Action & $[22, 21, 20, 24]$ & $[22, 21, 17, 10]$ \\
Layout & $[22, 21, 20, 25]$ & $[22, 21, 17, 10]$ \\
\bottomrule
\end{tabular}
\end{table}

\Cref{tab:taps} compares the selected taps against the four best-scoring band
layers and against the even spacing the image stream uses. The selected taps lead on
four of the five metrics, most clearly on text rendering; GenEval is the
exception. We did
not sweep it further, so the choice is supported empirically rather than
established as optimal.

\begin{table}[!h]
\centering
\small
\setlength{\tabcolsep}{6pt}
\renewcommand{\arraystretch}{1.15}
\caption{\textbf{Where the text-stream heads read.} Trained for 1{,}000 steps
and evaluated at 1 NFE.}
\label{tab:taps}
\begin{tabular}{@{}lccccc@{}}
\toprule
Text-stream taps & LongText-EN & LongText-CN & GenEval & DPG-Bench & OneIG \\
\midrule
Selected (\cref{tab:layers}) & \textbf{24.47} & \textbf{26.63} & 64.39 & \textbf{84.86} & \textbf{40.02} \\
Top four & 21.26 & 25.17 & 64.91 & 83.40 & 36.53 \\
Evenly spaced, $[5, 13, 21, 29]$ & 22.56 & 24.67 & \textbf{67.62} & 84.74 & 38.71 \\
\bottomrule
\end{tabular}
\end{table}

\section{Additional Results}
\label{sec:suppl:extra}

\begin{figure}[p]
\centering
\includegraphics[width=\linewidth]{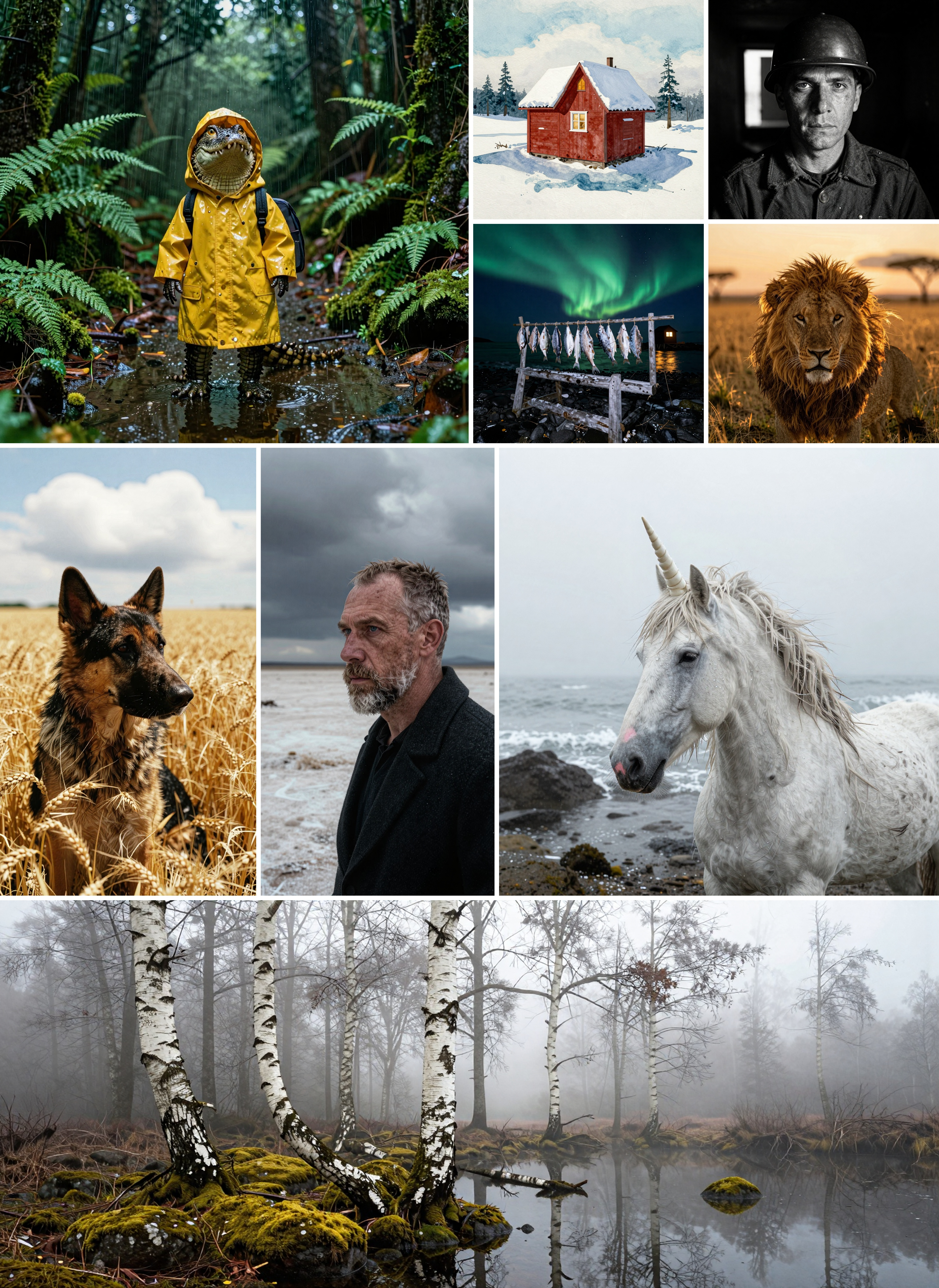}
\caption{Additional two-step samples from \ours, at $1024\times1024$.}
\label{fig:gal2}
\end{figure}

\begin{figure}[p]
\centering
\includegraphics[width=\linewidth]{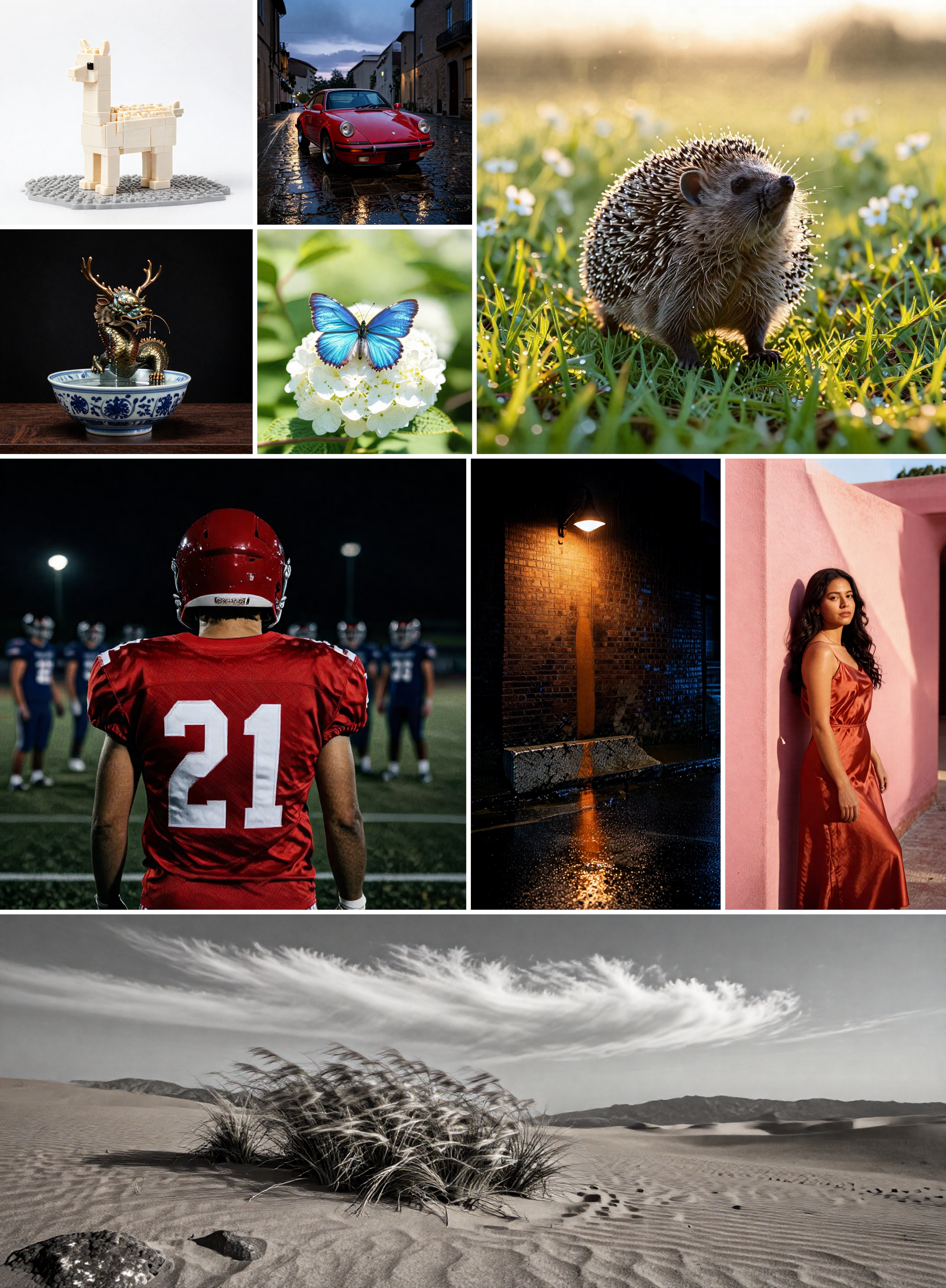}
\caption{Additional single-step samples from \ours, at $1024\times1024$.}
\label{fig:gal1}
\end{figure}

\cref{tab:oneig} breaks OneIG-Bench down by dimension and \cref{tab:dpg}
breaks DPG-Bench down by category. \cref{fig:gal2,fig:gal1} show further
two-step and single-step samples.

On the OneIG diversity dimension (\cref{tab:oneig}), \ours scores below both
baselines at one and two steps.
This suggests a trade-off between diversity and fidelity
\cite{lin2024lightning}.

\begin{table}[!h]
\centering
\small
\setlength{\tabcolsep}{4pt}
\caption{\textbf{Per-dimension breakdown on OneIG-Bench.} Results on the five
dimensions that form the overall score.}
\label{tab:oneig}
\begin{tabular}{lcccccccc}
\toprule
Method & Params & NFE & Overall & Alignment & Text & Diversity & Style & Reasoning \\
\midrule
\rowcolor{gray!12}
\ours           & 6B & 1 & \textbf{47.63} & \textbf{85.13} & \textbf{76.28} & 8.50 & \textbf{43.00} & \textbf{25.24} \\
Z-Image-Turbo   & 6B & 1 & 32.43 & 65.57 & 29.79 & 12.68 & 31.37 & 22.75 \\
TwinFlow        & 6B & 1 & 42.20 & 81.68 & 54.13 & \textbf{14.97} & 36.10 & 24.12 \\
\midrule
\rowcolor{gray!12}
\ours           & 6B & 2 & 51.45 & 85.41 & 91.34 & 9.63 & \textbf{44.20} & 26.68 \\
Z-Image-Turbo   & 6B & 2 & 50.94 & 83.92 & 92.85 & 13.85 & 36.25 & 27.86 \\
TwinFlow        & 6B & 2 & 51.38 & 85.26 & 90.61 & \textbf{15.49} & 38.03 & 27.52 \\
Z-Image-Turbo++ & 6B+6B & 2 & \textbf{52.50} & \textbf{85.65} & \textbf{97.09} & 11.75 & 37.87 & \textbf{30.12} \\
\midrule
Z-Image-Turbo   & 6B & 8 & 52.84 & 84.03 & 99.32 & 13.85 & 36.81 & 30.19 \\
\bottomrule
\end{tabular}
\end{table}

\begin{table}[!h]
\centering
\small
\setlength{\tabcolsep}{4pt}
\caption{\textbf{Per-category breakdown on DPG-Bench.} Results on the five
categories that form the overall score.}
\label{tab:dpg}
\begin{tabular}{lcccccccc}
\toprule
Method & Params & NFE & Overall & Entity & Attribute & Relation & Global & Other \\
\midrule
\rowcolor{gray!12}
\ours           & 6B & 1 & \textbf{86.48} & \textbf{92.36} & \textbf{89.62} & \textbf{94.74} & \textbf{84.19} & \textbf{88.00} \\
Z-Image-Turbo   & 6B & 1 & 77.23 & 84.85 & 83.08 & 91.03 & 76.90 & 82.80 \\
TwinFlow        & 6B & 1 & 85.17 & 90.99 & 88.56 & 94.43 & 83.89 & 87.20 \\
\midrule
\rowcolor{gray!12}
\ours           & 6B & 2 & \textbf{86.21} & \textbf{92.47} & 88.94 & 93.81 & 83.28 & 86.00 \\
Z-Image-Turbo   & 6B & 2 & 85.78 & 91.73 & 88.78 & 94.16 & \textbf{84.19} & \textbf{89.60} \\
TwinFlow        & 6B & 2 & 85.98 & 91.83 & 89.02 & 94.16 & 81.46 & 87.60 \\
Z-Image-Turbo++ & 6B+6B & 2 & 85.86 & 91.84 & \textbf{89.30} & \textbf{94.58} & 83.89 & \textbf{89.60} \\
\midrule
Z-Image-Turbo   & 6B & 8 & 84.86 & 89.59 & 90.14 & 92.16 & 91.29 & 88.68 \\
\bottomrule
\end{tabular}
\end{table}

Beyond the few-step students of Z-Image, we compare with two recent open-weight
models using their published two-step Turbo LoRAs: Krea 2 Turbo
\cite{krea2turbo,lvladikov2026krea2step} at 12B parameters and
Qwen-Image-2512 Turbo \cite{wu2025qwenimage,wuli2026qwen2step} at 20B.

\begin{figure}[H]
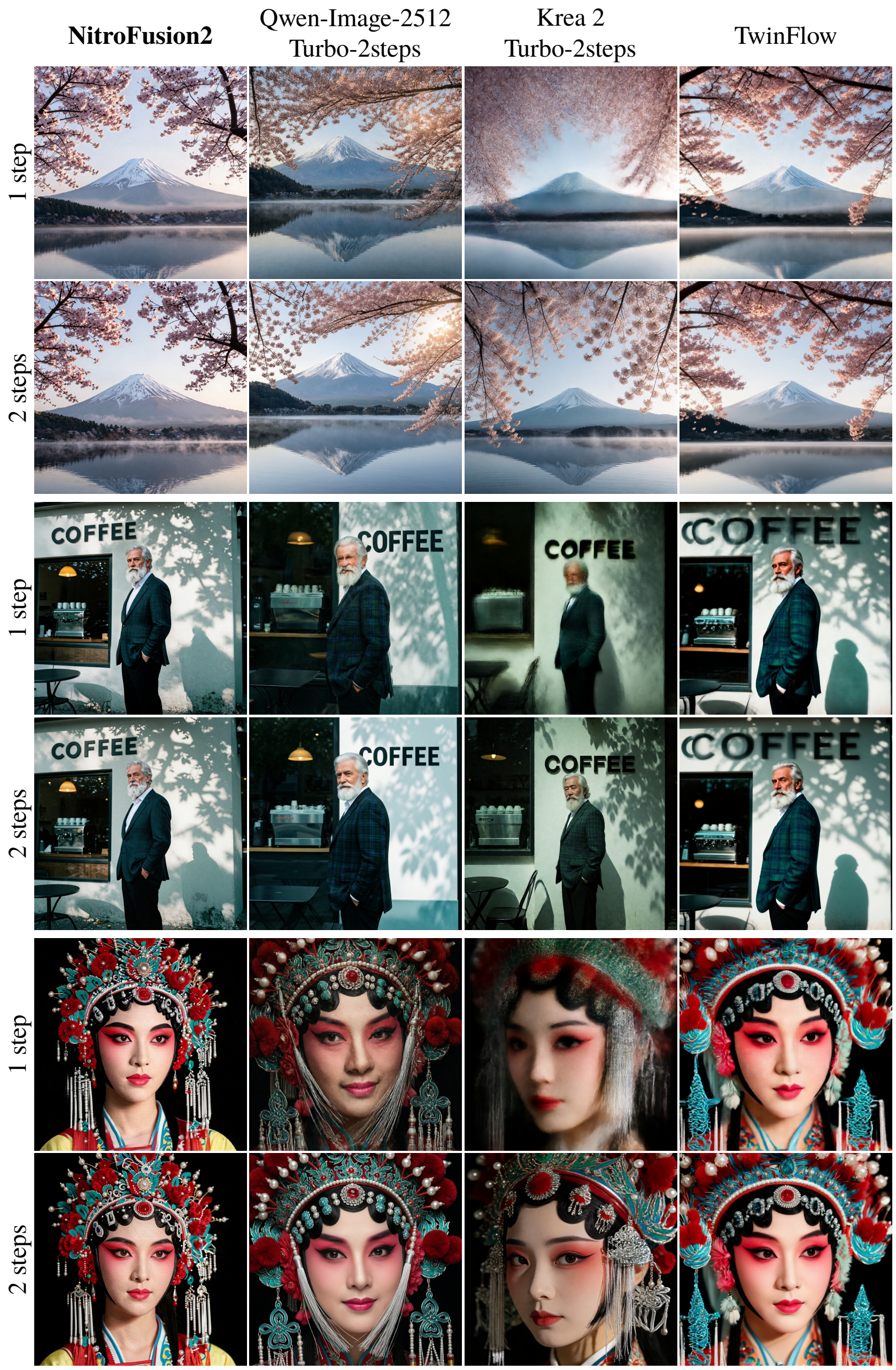

\centering
\newlength{\cmpvw}\setlength{\cmpvw}{0.188\linewidth}%
\newcommand{\cmpvcell}[3]{\includegraphics[width=\cmpvw]{figures/cmp4v/#1_#2_#3.pdf}}%
\newcommand{\cmpvhdr}[1]{\parbox[c][22pt][c]{\cmpvw}{\centering\figlbl #1\par}}%
\newcommand{\cmpvrow}[2]{%
  \rotatebox{90}{\makebox[\cmpvw]{\figlbl #2 step\ifnum#2>1 s\fi}} &
  \cmpvcell{#1}{nf2}{#2} & \cmpvcell{#1}{qwen2512}{#2} &
  \cmpvcell{#1}{krea2}{#2} & \cmpvcell{#1}{twinflow}{#2}}%
{\setlength{\tabcolsep}{0.5pt}\renewcommand{\arraystretch}{0}%
\begin{tabular}{@{}c cccc@{}}
  & \cmpvhdr{\textbf{\ours}} & \cmpvhdr{Qwen-Image-2512\\Turbo-2steps} &
  \cmpvhdr{Krea 2\\Turbo-2steps} & \cmpvhdr{TwinFlow} \\[3pt]
  \cmpvrow{fuji}{1} \\[1pt]
  \cmpvrow{fuji}{2} \\[3pt]
  \cmpvrow{coffee}{1} \\[1pt]
  \cmpvrow{coffee}{2} \\[3pt]
  \cmpvrow{opera}{1} \\[1pt]
  \cmpvrow{opera}{2} \\
\end{tabular}}

\caption{\textbf{Additional qualitative comparison at one and two steps.}}
\label{fig:compare4v}
\end{figure}

\Cref{fig:compare4v} shows the qualitative comparison. At one step, only \ours
produces consistently sharp outputs across all three examples, preserving fine
detail in the cherry-blossom branches, facial hair, and ornate headdresses. At
two steps, the differences become smaller, while \ours continues to produce
fine-grained detail with a 6B model, half the size of Krea 2 Turbo and less
than a third of Qwen-Image-2512 Turbo.
\Cref{tab:competitors} reports the quantitative results. At one step, \ours
leads four of the five benchmarks, trailing only on LongText-CN. At two steps,
Qwen-Image-2512 Turbo leads on text rendering and Krea 2 Turbo on GenEval,
while the three models are close on DPG-Bench and OneIG, leaving \ours
competitive with both larger models.

\vspace{-6mm}

\begin{table}[H]
\centering
\small
\setlength{\tabcolsep}{3pt}
\caption{\textbf{Comparison against recent open-weight models at one and two
steps.} Every row was sampled by us under identical settings. Parameter counts
are denoiser weights.}
\label{tab:competitors}
\begin{tabular}{lccccccc}
\toprule
& & & \multicolumn{2}{c}{Text rendering} & \multicolumn{3}{c}{Overall} \\
\cmidrule(lr){4-5}\cmidrule(lr){6-8}
Method & Params & NFE & LongText-EN & LongText-CN & GenEval & DPG-Bench & OneIG \\
\midrule
\rowcolor{gray!12}
\textbf{\ours} (ours)   & 6B  & 1 & \textbf{43.34} & 47.78 & \textbf{75.86} & \textbf{86.48} & \textbf{47.63} \\
Qwen-Image-turbo-2steps & 20B & 1 & 40.44 & \textbf{51.50} & 69.34 & 86.06 & 40.03 \\
Krea2-turbo-2steps      & 12B & 1 & 13.86 & 3.67 & 51.24 & 74.62 & 30.47 \\
TwinFlow                & 6B  & 1 & 29.18 & 36.85 & 66.34 & 85.17 & 42.20 \\
\midrule
\rowcolor{gray!12}
\textbf{\ours} (ours)   & 6B  & 2 & 76.03 & 81.95 & 74.46 & 86.21 & 51.45 \\
Qwen-Image-turbo-2steps & 20B & 2 & \textbf{89.42} & \textbf{93.00} & 75.28 & 86.26 & 50.79 \\
Krea2-turbo-2steps      & 12B & 2 & 86.78 & 50.28 & \textbf{85.52} & \textbf{86.67} & \textbf{52.59} \\
TwinFlow                & 6B  & 2 & 71.99 & 78.65 & 72.41 & 85.98 & 51.38 \\
\bottomrule
\end{tabular}
\end{table}

\clearpage
\section{Pseudo-code}
\label{sec:suppl:algo}

\begin{algorithm}[H]
\caption{One training step of \ours. ACD draws the cue and routes the heads
(lines 4--5). TSAD reads the text stream (line 9). Everything else is a standard
adversarial distillation step.}
\label{alg:train}
\begin{algorithmic}[1]
\Require Frozen MMDiT backbone $F$; student $G_\theta$; text-stream head groups
$\{D^{k}_{\mathrm{txt}}\}$, one per cue category; image-stream heads
$D_{\mathrm{img}}$; reference set $\mathcal D$ of pairs $(x,c_{\mathrm{full}})$
with original prompt $p$
\While{not converged}
  \State $(x,c_{\mathrm{full}},p)\sim\mathcal D$, \quad
         $z\sim\mathcal N(0,\mathbf I)$
  \State \textbf{Student output:} $\hat x\leftarrow G_\theta(z,1,p)$
  \State \textbf{ACD, draw the cue:} $k\sim\operatorname{Uniform}(\mathcal K(a))$,
         \quad $q\leftarrow$ field $k$ of $c_{\mathrm{full}}$
  \State \textbf{ACD, route the heads:} draw $D_{\mathrm{txt}}$ from the group $D^{k}_{\mathrm{txt}}$
  \State $t\sim\mathcal U$, \quad $\epsilon\sim\mathcal N(0,\mathbf I)$, \quad
         form $x_t$ and $\hat x_t$
  \State \textbf{One frozen pass per sample, both conditioned on $q$:}
  \State \quad $U_T,U_I\leftarrow F(x_t,q)$, \quad
         $\hat U_T,\hat U_I\leftarrow F(\hat x_t,q)$
  \State \textbf{TSAD, read the text stream:}
         $\ell^{\mathrm{text}}_{q}\leftarrow D_{\mathrm{txt}}(U_T)$, \quad
         $\hat\ell^{\mathrm{text}}_{q}\leftarrow D_{\mathrm{txt}}(\hat U_T)$
  \State \textbf{Image stream, as usual:}
         $\ell^{\mathrm{img}}_{q}\leftarrow D_{\mathrm{img}}(U_I)$, \quad
         $\hat\ell^{\mathrm{img}}_{q}\leftarrow D_{\mathrm{img}}(\hat U_I)$
  \State \textbf{Update the heads:} descend $\mathcal L_{\mathrm D}$ on
         $D_{\mathrm{txt}},D_{\mathrm{img}}$ with $\hat x_t$ detached
  \State \textbf{Update the student:} descend $\mathcal L_{\mathrm G}=
         (1-\lambda_{\mathrm{img}})\mathcal L_{\mathrm{text\text{-}adv}}
         +\lambda_{\mathrm{img}}\mathcal L_{\mathrm{img\text{-}adv}}
         +\lambda_{\mathrm{fm}}\mathcal L_{\mathrm{fm}}$ on $G_\theta$
\EndWhile
\State \Return trained student $G_\theta$
\end{algorithmic}
\end{algorithm}

\end{document}